\documentclass[10pt,twocolumn,letterpaper]{article}

\usepackage[]{cvpr}      %

\usepackage[dvipsnames,table,xcdraw]{xcolor}
\usepackage{ulem}

\usepackage{pifont}
\usepackage{makecell}
\usepackage{dsfont}
\usepackage{multirow}
\usepackage{listings}
\usepackage{algorithm}
\usepackage{algpseudocode}
\usepackage{tabularx}

\newcommand{\ours}[0]{VISCO}
\newcommand{\ourscore}[0]{VISCore}
\newcommand{\ourscritic}[0]{\textsc{LookBack}}
\newcommand{\mypar}[1]{\vspace{.2em}\noindent\textbf{#1}}

\definecolor{amber}{rgb}{1.0, 0.49, 0.0}
\definecolor{darkred}{rgb}{0.75, 0.0, 0.0}
\definecolor{mygreen}{HTML}{00A64F}

\definecolor{cvprblue}{rgb}{0.21,0.49,0.74}
\usepackage[pagebackref,breaklinks,colorlinks,citecolor=cvprblue]{hyperref}

\def\paperID{9260} %
\def\confName{CVPR}
\def\confYear{2025}

\title{\ours{}: Benchmarking Fine-Grained Critique and Correction Towards Self-Improvement in Visual Reasoning\vspace{-12pt}}

\author{Xueqing Wu$^{*1}$, Yuheng Ding$^{*1}$, Bingxuan Li$^1$, Pan Lu$^2$, Da Yin$^1$, Kai-Wei Chang$^1$, Nanyun Peng$^1$ \\
$^1$ University of California, Los-Angeles~~~~$^2$ Stanford \\
\url{https://visco-benchmark.github.io/}}

\begin{document}
\maketitle

\makeatletter
\renewcommand{\@makefnmark}{$^*$}%
\footnotetext{Equal contribution. Contact: \texttt{xueqing.wu@cs.ucla.edu}}
\makeatother

\begin{abstract}
\vspace{-6pt}The ability of large vision-language models (LVLMs) to critique and correct their reasoning is an essential building block towards their self-improvement. However, a systematic analysis of such capabilities in LVLMs is still lacking. We propose \ours{}, the first benchmark to extensively analyze the fine-grained critique and correction capabilities of LVLMs. Compared to existing work that uses a single scalar value to critique the entire reasoning \citep{mllm_judge}, \ours{} features \textbf{dense} and \textbf{fine-grained} critique, requiring LVLMs to evaluate the correctness of each step in the chain-of-thought and provide natural language explanations to support their judgments. Extensive evaluation of 24 LVLMs demonstrates that human-written critiques significantly enhance the performance after correction, showcasing the potential of the self-improvement strategy. However, the model-generated critiques are less helpful and sometimes detrimental to the performance, suggesting that critique is the crucial bottleneck. We identified three common patterns in critique failures: failure to critique visual perception, reluctance to ``say no'', and exaggerated assumption of error propagation. To address these issues, we propose an effective \ourscritic{} strategy that revisits the image to verify each piece of information in the initial reasoning. \ourscritic{} significantly improves critique and correction performance by up to 13.5\%.
\end{abstract}
    
\vspace{-1em}
\section{Introduction}
\label{sec:intro}

With recent advances, large vision-language models (LVLMs) have unlocked strong reasoning capabilities \citep{gpt4,gpt4v}, enabling them to solve complex problems in math \citep{mathvista,mathvision} and science \citep{mmmu,scemqa}. Typically, LVLMs adopt the \textbf{chain-of-thought} (CoT) approach \citep{cot}, where the model generates intermediate reasoning steps before reaching the final answer. However, even with the CoT approach, LVLMs remain prone to hallucination \citep{pope,hallusionbench,gunjal2024detecting} and reasoning errors \citep{campbellunderstanding}, especially when dealing with compositional concepts like counting \citep{tallyqa,paiss2023teaching} and identifying spatial relationship \citep{vsr,embspatial,kamath2023s}. Such issues raise concern about the trustworthiness of visual reasoning and highlights the need for more reliable visual reasoning.

\begin{figure}[!t]
    \centering
\includegraphics[width=\linewidth]{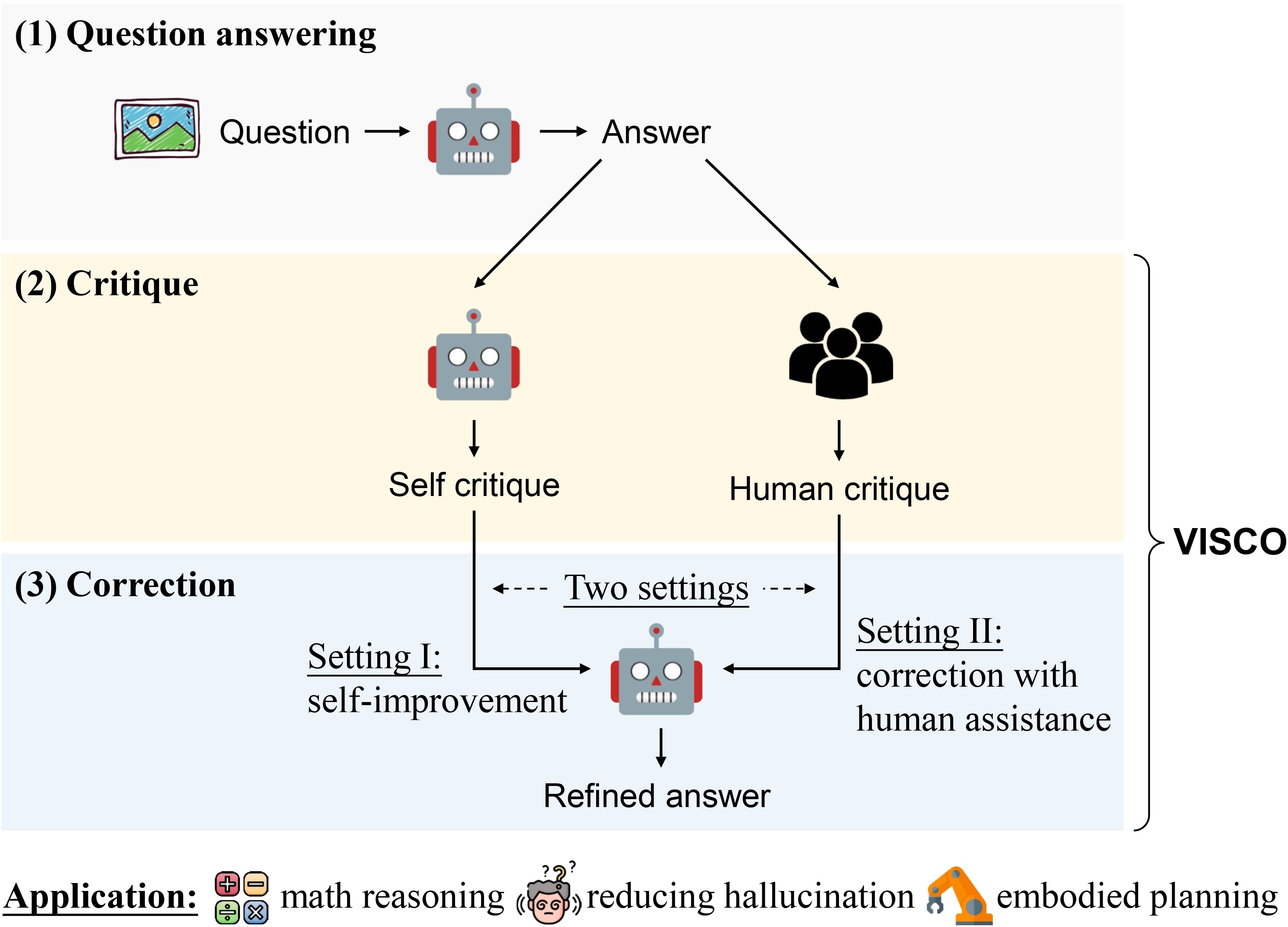}
\vspace{-1.8em}
    \caption{\textbf{Evaluation settings of \ours{}.} We evaluate the self-improvement setting where LVLMs correct their own answers based on self-generated critique, and a human-assisted setting where correction is based on human-written critique.}
    \label{fig:teaser_small}
    \vspace{-16pt}
\end{figure}

\begin{figure*}[!t]
    \centering
\includegraphics[width=\linewidth]{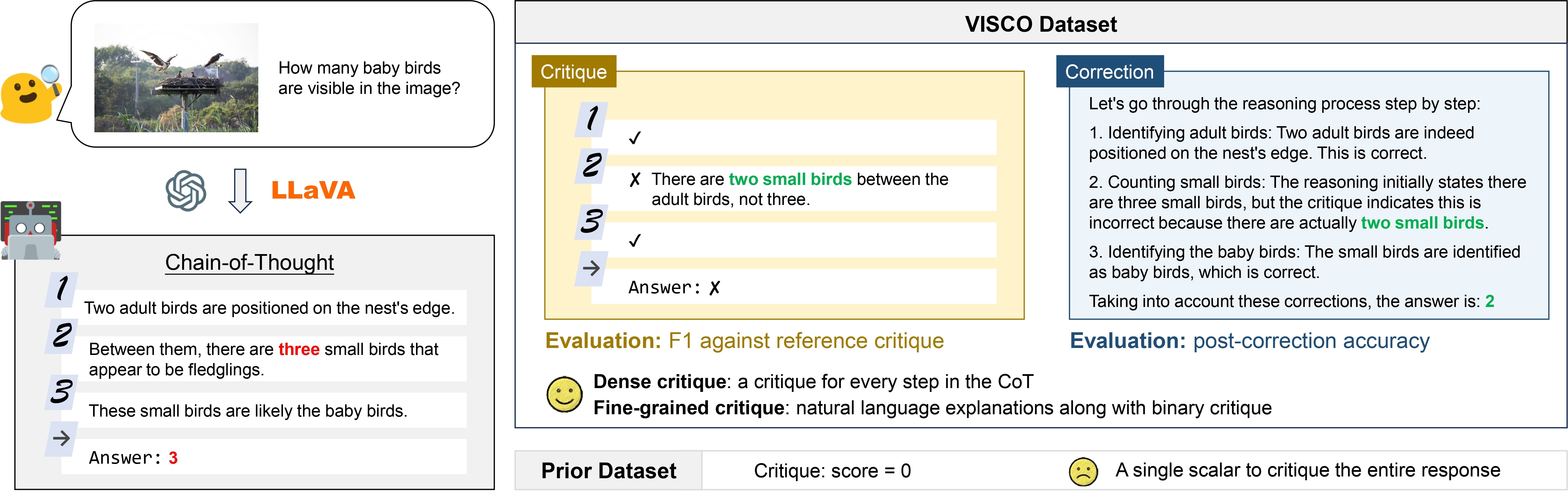}
\vspace{-1.7em}
    \caption{\textbf{Overview of \ours{} task, data and evaluation.} \ours{} dataset features step-wise critique with natural language explanations.}
    \label{fig:teaser}
    \vspace{-12pt}
\end{figure*}

Given the difficulty of generating completely accurate reasoning in a single pass, an alternative strategy is \textbf{self-improvement}, where LVLMs \textit{critique} their initial reasoning and make \textit{corrections} accordingly to enhance performance.
Application to text-only tasks like code generation \citep{madaan2024self,chen2023teaching} and multi-hop question answering \citep{reflexion} has demonstrated the effectiveness of self-improvement.
For vision-language tasks, the self-improvement strategy has also been applied to areas including video understanding \citep{mahmood2024vurf} and region-of-interests identification \citep{yu2024attention}. %
However, a comprehensive study of LVLMs' self-improvement is still lacking.
While some work evaluates the critique capability and enhances it through specialized training, they use LVLMs as evaluators \citep{prometheus} or reward models \citep{llavacritic} merely to assess the output quality, rather than for further self-correction and self-improvement.

To conduct a systematic analysis of the \textbf{critique} and \textbf{correction} capabilities of LVLMs, we present \textbf{\ours{}}, the first benchmark for VIsual Self-Critique and cOrrection. As shown in Figure \ref{fig:teaser_small}, we evaluate critique and correction as two stages in the self-improvement pipeline, and additionally evaluate a setting where LVLMs correct model responses based on high-quality human-annotated critique. We evaluate a wide range of 18 datasets across 8 tasks, divided into two main categories: \underline{reasoning} tasks such as math reasoning, and \underline{perception} tasks like reducing hallucination. We sample chain-of-thoughts from 7 LVLMs and have expert human annotators produce ground-truth critiques. While existing work typically uses a single scalar to critique the entire model response \citep{mllm_judge}, research in education shows that detailed and specific feedback is more effective for student learning and error correction \citep{yunus2020written,lipnevich2021review}. Therefore, we collect \textbf{dense and fine-grained} critique, assigning a binary correctness label for \textit{each step} in the chain-of-thought, followed by a natural language explanation if the step is incorrect. Our dataset consists of 1645 question-answer pairs, including 5604 \textit{step-wise} annotations. An example of the data and task is shown in Figure \ref{fig:teaser}.

We extensively evaluate 24 LVLMs, including open-source models like LLaVA \citep{llava} and Molmo \citep{molmo} and proprietary models like GPT-4o and Claude-3.5. Our results show that human-written critiques significantly contribute to the correction task, enabling LVLMs to correct up to 76\% of the errors. This demonstrates the great potential of self-improvement. Despite these improvements, the benefits decrease substantially when using critiques generated by LVLMs, showing that \textbf{effective critique is the key bottleneck}.
In the critique task, we observe that non-trivial, above-random critique capabilities typically emerge in LVLMs with $>$70B parameters.
However, even the leading LVLMs still show a significant gap compared to human experts. Specifically, we identify three common failure patterns in the critique task: (1) \uline{Failure to critique visual perception}, where LVLMs have more difficulty critiquing visual perception than verbal reasoning; (2) \uline{Reluctance to ``say no''}, where LVLMs are biased towards judging the input CoT as correct; (3) \uline{Exaggerated assumption of error propagation}, where LVLMs overestimate the likelihood that errors made in earlier steps of the CoT will propagate to later steps.

To address these issues, we propose a simple but effective critique strategy: \textbf{\ourscritic{}}. \ourscritic{} first identifies information in the CoT that needs to be verified against the image. It then revisits the image to explicitly verify each piece of information, thus mitigating the difficulty in critiquing visual perception. The inconsistency revealed between the CoT and the image is naturally leveraged to perform critique, further addressing the reluctance to ``say no''. Experiments across four LVLMs show that \ourscritic{} significantly enhances critique performance by up to 13.5\% with even greater gains (22.2\%) in perception tasks. When applied to the correction task, critiques produced by \ourscritic{} further boost the performance by up to 11.5\%.

In summary, our contributions are three-fold: (1) We propose \ours{}, the first benchmark to evaluate critique and correction in visual reasoning, featuring densely annotated step-wise critique and natural language explanations; (2) We conduct an extensive evaluation of 24 LVLMs and summarize three failure patterns for the critique task to inspire future research; (3) We propose \ourscritic{}, an improved baseline that significantly improves both critique and correction by up to 13.5\%.

\section{Related Work}

\begin{table}[!htbp]
    \centering
    \footnotesize
\setlength{\tabcolsep}{1.6pt}
\begin{tabular}{l|ccccc}
\toprule
& \textbf{\footnotesize \makecell{Input \\ Modality}} & \textbf{\footnotesize \makecell{Critique \\ Format}} & \textbf{\footnotesize \makecell{Dense \\ Critique}} & \textbf{\footnotesize \makecell{Human \\ Annotation}} \\
\midrule
CriticBench \citep{criticbench} & T & Scalar & \color{red}\ding{55} & \color{mygreen}\ding{51} \\
MetaCritique \citep{metacritique} & T & Scalar & \color{mygreen}\ding{51} & \color{red}\ding{55} \\
CriticEval \citep{criticeval} & T & Scalar+NL & \color{red}\ding{55} & \color{mygreen}\ding{51} \\
\midrule
MLLM-as-a-Judge \citep{mllm_judge} & T+V & Scalar & \color{red}\ding{55} & \color{mygreen}\ding{51} \\
Perception Collection \citep{prometheus} & T+V & Scalar+NL & \color{red}\ding{55} & \color{red}\ding{55} \\
\textbf{\ours{} (Ours)} & \textbf{T+V} & \textbf{Scalar+NL} & \textbf{\color{mygreen}\ding{51}} & \textbf{\color{mygreen}\ding{51}} \\
\bottomrule
    \end{tabular}
    \vspace{-1em}
    \caption{\textbf{Comparison against existing datasets for critique.} For input modality, T represents text and V represents vision. For critique format, NL represents natural language.}
    \label{tab:related}
    \vspace{-15pt}
\end{table}

\mypar{Visual reasoning.} Large-scale pre-training on extensive texts and images has equipped LVLMs with strong reasoning capabilities \citep{gpt4v,llava}. Various work has evaluated their capabilities across different dimensions, such as mathematical \citep{mathvista, mathvision}, scientific \citep{mmmu, scemqa}, and spatial relationship reasoning \citep{vsr, embspatial, kamath2023s}. A common reasoning strategy is the chain-of-thought approach, where LLMs or LVLMs first generate reasoning rationales before providing final answers \citep{cot,chen2023measuring}. While most work focuses on textual rationales, some work also explores other forms of rationale such as region of interests \citep{shao2024visual,liu2024chain} or sketches \citep{visualsketchpad}. In this work, we focus on reasoning in text format.

\mypar{Self-improvement.} The self-improvement strategy, also termed self-refinement, has been successfully applied to LLMs and LVLMs. By iteratively critiquing and correcting their responses, LLMs have shown improved performance in tasks such as code generation \citep{madaan2024self,chen2023teaching}, multi-hop question answering \citep{reflexion}, and agent development \citep{voyager}. For visual reasoning, this strategy has been employed in text-to-image generation \citep{yang2023idea2img}, video understanding \citep{mahmood2024vurf}, region-of-interest identification \citep{yu2024attention}, and instruction data generation \citep{wang2024enhancing}. However, a comprehensive analysis of LVLMs self-improvement capabilities is still lacking.

\mypar{Datasets for critique and correction.}  The most relevant prior work includes text-only benchmarks that evaluate the critique and correction abilities of LLMs \citep{luocritique,criticbench,metacritique,criticeval}. In the multimodal domain, \citet{mllm_judge} and \citet{prometheus} focus on using LVLMs as evaluators to critique the instruction following performance in model responses. However, they do not focus on critiquing the correctness of model responses or consider the self-correction step. Table \ref{tab:related} highlights other limitations of existing work, including the lack of dense critique, reliance on scalar-format critique without detailed natural language explanations, and the lack of human-annotated critique.

\section{\ours{} Benchmark}

We present \ours{}, the first benchmark to evaluate the critique and correction capabilities of LVLMs. \ours{} consists of 1645 pairs of questions and LVLM-generated answers, each densely annotated with step-wise binary labels and natural language explanations. The dataset spans 18 datasets and 8 tasks across two main categories: (1) reasoning tasks, such as math and science reasoning, and (2) perception tasks, such as text recognition, spatial relationship understanding, and prevention of hallucination. In the following sections, we first provide the mathematical formulations of our task and evaluation in Section \ref{sec:task_formulation}. Next, we describe the data collection and annotation process in Section \ref{sec:dataset_construction}. Finally, we present the dataset statistics in Section \ref{sec:dataset_statistics}.

\subsection{Task Overview}
\label{sec:task_formulation}

Given images $I$ and questions $q$ from existing datasets, we use LVLMs $\mathcal{M}$ to sample model responses. Following the CoT approach, a model response contains a chain-of-thought $\mathbf{s}_\mathcal{M} = [s_1, \ldots, s_N]$ with $N$ reasoning steps, leading to the final answer $a_\mathcal{M}$. Additionally, we have the ground truth answers $a_\text{gt}$ from the original datasets.
Using the tuple $(I, q, a_\text{gt}, \mathbf{s}_\mathcal{M}, a_\mathcal{M})$, we evaluate two tasks, critique and correction.

\subsubsection{Critique}

Given input $I,q$, the critique task produces a critique $\mathcal{C}$ to judge the \textbf{correctness} of model response $(\mathbf{s}_\mathcal{M}, a_\mathcal{M})$. The judgment is based on the accuracy of factual information and logical calculations in the response. 
Motivated by research in the education field \citep{yunus2020written,lipnevich2021review}, we propose the use of \textbf{dense} and \textbf{fine-grained} critique, including: %
\begin{enumerate}
    \item A binary critique $\mathcal{C}_{s,i}$ for each intermediate step $s_i$; %
    \item A natural language explanation $\mathcal{C}_{e,i}$ to explain the binary critique $\mathcal{C}_{s,i}$ for step $s_i$;
    \item And finally, a binary critique $\mathcal{C}_a$ for the final answer $a_\mathcal{M}$.
\end{enumerate}
Therefore, the output of the critique task is a hierarchical structure $\mathcal{C} = (\{(\mathcal{C}_{s,i}, \mathcal{C}_{e,i})\}_{i=1}^N, \mathcal{C}_a)$.

\mypar{Hierarchical evaluation.} Given the complex structure of critique $\mathcal{C}$, we first evaluate the critique at each granularity level and then combine the results into the final metric.
We adopt F1 as our base metric and consider \textit{incorrect} as the positive class, thus calculating the F1 of correctly identified errors. We calculate F1 scores at three hierarchical levels: $\text{F1}_a$ for answer critique $\mathcal{C}_a$, $\text{F1}_s$ for step-wise critique $\mathcal{C}_s$, and $\text{F1}_e$ for critique explanation $\mathcal{C}_e$. While it is straightforward to calculate F1 for binary critique $\mathcal{C}_a$ and $\mathcal{C}_s$, measuring F1 for natural language explanation $\mathcal{C}_e$ poses a challenge. Therefore, we adopt an LLM-assisted evaluation approach described below.

\mypar{LLM-assisted evaluation of natural language explanation.} Since each explanation $\mathcal{C}_{e,i}$ aims to justify the binary step-wise critique $\mathcal{C}_{s,i}$, we adjust the calculation of step-wise F1$_s$ to derive F1$_e$. Specifically, we recalculate the \textbf{true positives} as the number of steps $s_i$ that are (1) correctly judged as incorrect, and (2) have their errors correctly explained in $\mathcal{C}_{e,i}$. This evaluation only considers explanations for incorrect steps, as explanations for correct steps can be as trivial as reiterating the correct reasoning.
We use LLMs to assess the accuracy of critique explanations by comparing them to human-annotated explanations. While minor deviations or omissions are tolerated, the model-generated critique must address the core error without introducing more factual or logical errors to be considered as correct. As shown by existing work \citep{geval}, LLMs can reliably evaluate free-form text given well-defined evaluation criteria. %

\begin{figure}
        \centering
        \includegraphics[width=\linewidth]{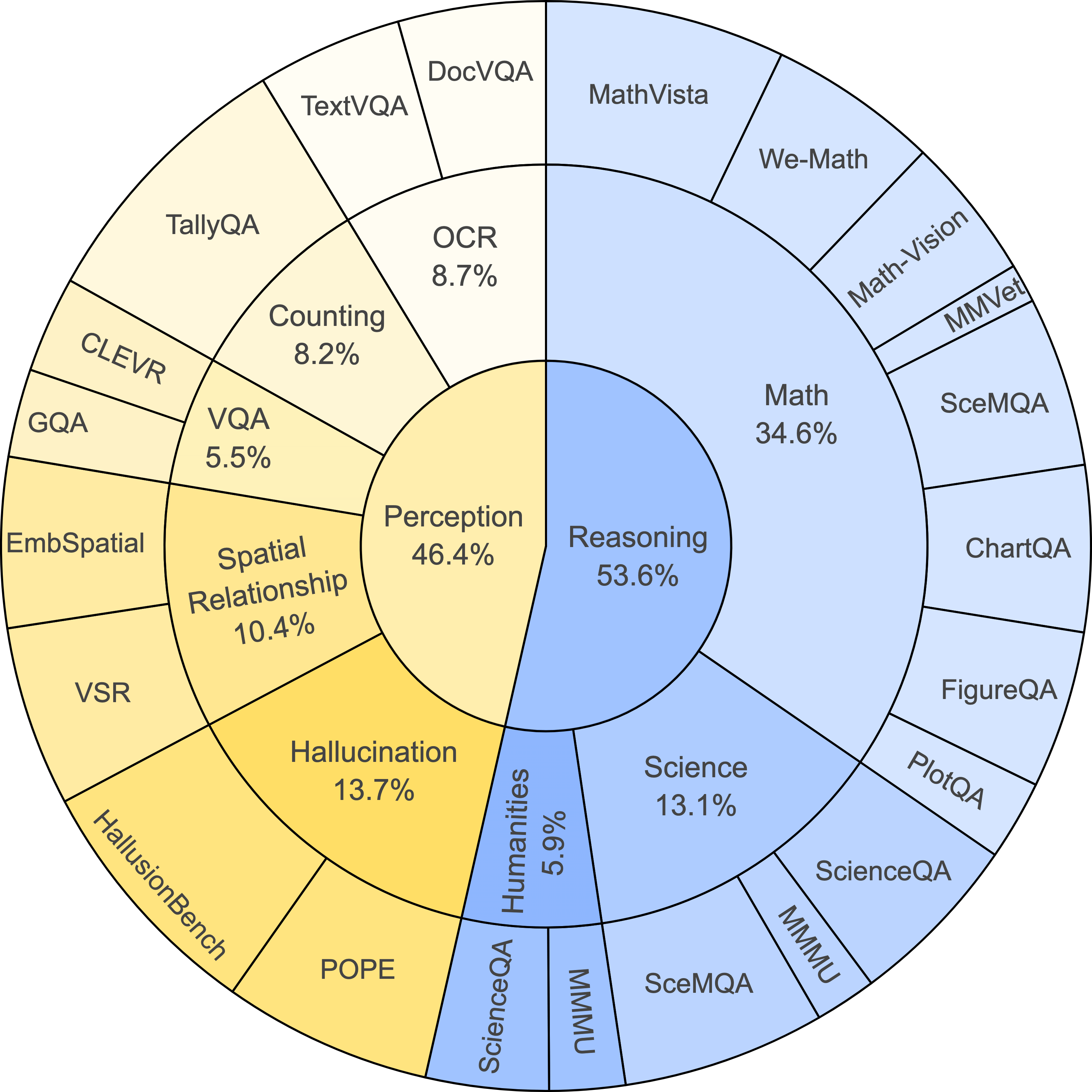} %
        \vspace{-1.8em}
        \caption{\textbf{Categories, tasks and datasets distribution of \ours{}.}}
        \label{fig:statistics}
        \vspace{-16pt}
\end{figure}

\mypar{\ourscore{}.} Eventually, we combine the three scores F1$_a$, F1$_s$ and F1$_e$ into a holistic score \ourscore{}. Due to the stricter criteria for explanation correctness, its F1 score F1$_e$ is naturally of a lower scale than F1$_a$ and F1$_s$. Therefore, a key challenge lies in comprehensively considering all three scores without neglecting the information contained in scores with lower scales. To penalize models that ``guess'' accurate binary critique but fail to provide effective explanations, we calculate our final metric \ourscore{} as the geometric average of the three F1 scores, formally defined as:
$$\text{\ourscore{}} = \left(\text{F1}_a \times \text{F1}_s \times \text{F1}_e\right)^\frac{1}{3}$$
More details for \ourscore{} design are in Appendix \ref{sec:metric}.

\subsubsection{Correction}

Based on $I$, $q$ and model response $(\mathbf{s}_\mathcal{M}, a_\mathcal{M})$, the correction task takes critique $\mathcal{C}$ as input to generate a refined answer $a'$. As shown in Figure \ref{fig:teaser_small}, the critique $\mathcal{C}$ can be generated by either LVLMs or human experts.

\begin{table}[htbp]
\setlength{\tabcolsep}{4pt}
        \centering
        \begin{tabular}{l|r|rrr}
            \toprule
 & Total & \begin{tabular}{@{}r@{}}\footnotesize Outcome\vspace{-.4em} \\ \footnotesize Error \end{tabular} & \begin{tabular}{@{}r@{}}\footnotesize Process\vspace{-.4em} \\ \footnotesize Error \end{tabular} & \begin{tabular}{@{}r@{}}\footnotesize No\vspace{-.4em} \\ \footnotesize Error \end{tabular} \\
\midrule
 Questions & 1645 & 502 & 179 & 964 \\
~~- Incorrect answers & 502 & 502 & 0 & 0 \\
~~- Correct answers & 1143 & 0 & 179 & 964 \\
 Unique images  & 1587& 494 & 178 & 940 \\
\midrule
 Steps per question & 3.4 & 3.4 & 3.6 & 3.4 \\
 Steps in total & 5604 & 1713 & 650 & 3241 \\
~~- Incorrect steps & 1346 & 1046 & 300 & 0 \\
~~- Correct steps & 4258 & 667 & 350 & 3241 \\
 \midrule
 Average question length & 33.5 & 34.7 & 35.5 & 32.5 \\
 Average step length & 19.4 & 20.9 & 20.2 & 18.5 \\
 Average critique length & 17.0 & 17.2 & 16.1 & - \\
            \bottomrule
        \end{tabular}
        \vspace{-.6em}
        \caption{\textbf{Statistics of \ours{}.} The average length of questions, steps and critique are evaluated by the number of words.}
        \label{tab:statistics}
        \vspace{-15pt}
\end{table}

\mypar{Evaluation.} We first evaluate the accuracy of the refined answer $a'$ by comparing it against the ground truth $a_\text{gt}$. We then calculate \textbf{positive correction ratio} (PCR) representing how many errors are successfully corrected:
$$\text{PCR} = \underset{Acc(a_\text{gt}, a_\mathcal{M}) = 0}{\text{Average}} Acc(a_\text{gt}, a')$$
To prevent models from cheating the score by always shifting answers, we also calculate the \textbf{negative correction ratio} (NCR) evaluating how many correct answers are altered into incorrect ones:
$$\text{NCR} = \underset{Acc(a_\text{gt}, a_\mathcal{M}) = 1}{\text{Average}} \left(1 -Acc(a_\text{gt}, a')\right)$$
and report the \textbf{correction gain} $= \text{PCR}-\text{NCR}$ as the final metric. Detailed illustration of correction gain calculation is in Appendix \ref{sec:metric}.

\subsection{Dataset Construction}
\label{sec:dataset_construction}

The construction of \ours{} dataset consists of four steps: task input collection, response collection, response filtering, and critique collection. More examples of each step and the final dataset are in Appendix \ref{sec:dataset_examples}.

\mypar{Task input collection.} We collect input images $I$, questions $q$ and ground truth answers $a_\text{gt}$ from existing VQA datasets. We select a diverse range of \textbf{8 tasks} and \textbf{18 datasets}, divided into two categories: \uline{reasoning} tasks such as math reasoning \citep{mathvista,mathvision} and scientific reasoning \citep{mmmu,scemqa}, and \uline{perception} tasks such as hallucination \citep{pope,hallusionbench}, identifying spatial relationship \citep{vsr,embspatial}, and text recognition \citep{textvqa,docvqa}. Figure \ref{fig:statistics} shows a comprehensive visualization of task and dataset distribution.

\mypar{Response collection.}
We sample responses from 7 LVLMs of 3 families: InternVL2 8/26/40B, LLaVA-v1.6 7/13/34B, and OpenAI GPT-4o. We prompt LVLMs to first generate a CoT in a short paragraph and then generate the final answer. We then split the CoT paragraph into sentences, each representing a reasoning step $s_i$. To simplify the task, we limit the total number of reasoning steps $N$ within 5 and remove responses with more than 5 sentences in the CoT.

\mypar{Response filtering.} To evaluate whether LVLMs can correctly identify errors, we focus on two types of errors, \textit{outcome errors} and \textit{process errors}. Outcome errors refer to incorrect responses with incorrect final answers, and process errors refer to responses that reach the correct final answers but contain errors in the middle of its CoT. We filter out responses with outcome errors by comparing model-generated responses $a_\mathcal{M}$ against the ground truth $a_\text{gt}$ and select the mismatching data points. The matching results also serve as a preliminary annotation of answer critique $\mathcal{C}_a$, which will be further verified by human annotators. Within responses with correct final answers, we ask human annotators to manually filter out responses with process errors. We identify 179 responses with process errors from a total of 1143 responses. The responses with outcome or process errors will be further densely annotated by three annotators.

\mypar{Critique collection.} We recruit three expert annotators (computer science students with relevant AI experience) to annotate critique $\mathcal{C}$. The first 30 examples are used for training the annotators and aligning annotation criteria; these are not included in the final dataset. During the official annotation process, annotators first verify the automatically generated answer critique $\mathcal{C}_a$, and then assess the correctness of each step $s_i$, resulting in annotations of $\mathcal{C}_s$. This annotation achieves a substantial inter-annotator agreement of 64.76 Cohen's Kappa. For disagreed data points, we hold discussion sections every 50 samples where the annotators review each others' annotations and settle disagreements. Finally, we ask each annotator to write a natural language explanation for each incorrect step to collect $\mathcal{C}_e$, resulting in three references for each step. When evaluating model-generated critique explanation, we compare it against each of the three annotations and take the average score, leading to more accurate evaluation.

\subsection{Dataset Statistics}
\label{sec:dataset_statistics}

Detailed statistics of \ours{} are shown in Table \ref{tab:statistics}. In addition to the overall statistics, we show the breakdown of the subsets with outcome errors, process errors, and no errors respectively. In total, we collect 1645 data with 5604 steps, where 30\% data have outcome errors and 11\% have process errors. The distribution by categories (reasoning v.s. perception), tasks, and datasets are in Figure \ref{fig:statistics}.

\section{Experiments}
\label{sec:experiments}

In this section, we first evaluate the overall critique and correction capabilities of LVLMs and validate the importance of dense and fine-grained critique. While LVLMs in general perform well in the correction task, we find that critique remains a crucial bottleneck for effective self-improvement. We then conduct an in-depth analysis of the critique task and identify three common patterns of critique failure. These identified error patterns guide the design of our proposed \ourscritic{} method as in Section \ref{sec:ours_method}, and we hope they can inspire further research.%

\definecolor{pos}{RGB}{0,0,0} %
\definecolor{neg}{RGB}{240,0,0}

\begin{table}[!htbp]
    \centering
\begin{tabular}{ll|r|rr}
    \toprule
\multirow{2}{*}{Model} & \multirow{2}{*}{Size} & \multirow{2}{*}{Critique} & \multicolumn{2}{c}{Correction} \\
& & & $\mathcal{C}_\text{Human}$ & $\mathcal{C}_\text{Self}$ \\
\midrule
\multicolumn{5}{c}{\textit{Small-size ($\sim$7B) Open LVLMs}} \\
\midrule
DeepSeek-VL & 7B & 7.5 & \color{pos}2.3 & \color{neg}-17.9 \\
LLaVA-v1.6 & 7B & \underline{21.8} & \color{pos}40.3 & \color{neg}-8.7 \\
LLaVA-OV & 7B & 7.5 & \color{pos}22.4 & \color{neg}-9.1 \\
Qwen2-VL & 7B & 21.7 & \color{pos}50.8 & \color{pos}\underline{5.5} \\
Molmo & 7B & 13.4 & \color{pos}49.1 & \color{pos}1.8 \\
InternVL2 & 8B & \textbf{23.3} & \color{pos}\underline{52.7} & \color{pos}5.4 \\
MiniCPM-V2.6 & 8B & 13.1 & \color{pos}\textbf{53.0} & \color{pos}\textbf{7.8} \\
\midrule
\multicolumn{5}{c}{\textit{Medium-size (10$\sim$40B) Open LVLMs}} \\
\midrule
Llama-3.2 & 11B & 11.4 & \color{pos}34.8 & \color{neg}-11.7 \\
LLaVA-v1.6 & 13B & 21.0 & \color{pos}40.2 & \color{neg}-7.2 \\
InternVL2 & 26B & \underline{25.2} & \color{pos}\textbf{59.3} & \color{pos}\underline{6.0} \\
LLaVA-v1.6 & 34B & 11.1 & \color{pos}39.2 & \color{neg}-1.7 \\
InternVL2 & 40B & \textbf{28.5} & \color{pos}\underline{47.6} & \color{pos}\textbf{9.9} \\
\midrule
\multicolumn{5}{c}{\textit{Large-size ($>$70B) Open LVLMs}} \\
\midrule
LLaVA-OV & 72B & 35.3 & \color{pos}33.4 & \color{neg}-10.2 \\
Qwen2-VL & 72B & \textbf{37.4} & \color{pos}31.5 & \color{neg}-2.1 \\
NVLM & 72B & 33.1 & \color{pos}42.2 & \color{pos}1.7 \\
Molmo & 72B & 35.6 & \color{pos}53.1 & \color{pos}1.4 \\
InternVL2 & 76B & 26.4 & \color{pos}\textbf{72.7} & \color{pos}\textbf{11.7} \\
Llama-3.2 & 90B & \underline{36.4} & \color{pos}\underline{66.4} & \color{pos}\underline{4.4} \\
\midrule
\multicolumn{5}{c}{\textit{Critique LVLMs}} \\
\midrule
Prometheus-Vision & 7B & 17.7 & n/a$^\dagger$ & n/a$^\dagger$ \\
LLaVA-Critic & 7B & \underline{20.0} & \color{pos}\underline{19.3} & \color{neg}\underline{-11.4} \\
Prometheus-Vision & 13B & 19.3 & n/a$^\dagger$ & n/a$^\dagger$ \\
LLaVA-Critic & 72B & \textbf{42.6} & \color{pos}\textbf{58.9} & \color{pos}\textbf{15.4} \\
\midrule
\multicolumn{5}{c}{\textit{Proprietary LVLMs}} \\
\midrule
Gemini-1.5-Pro & * & 45.0 & \color{pos}\textbf{78.0} & \color{pos}24.9 \\
Claude-3.5-Sonnet & * & \underline{51.3} & \color{pos}73.7 & \color{pos}\underline{25.6} \\
GPT-4o & * & \textbf{52.4} & \color{pos}\underline{76.2} & \color{pos}\textbf{28.8} \\
\midrule
\rowcolor[RGB]{240,240,240} \multicolumn{2}{l|}{Human$^\ddagger$} & 86.5 & - & - \\
\bottomrule
    \end{tabular}
    \vspace{-.7em}
    \caption{\textbf{Leaderboard of \ours{}.} We report \ourscore{} for critique and correction gain for correction. For the correction task, we evaluate model performance with either human-annotated critique ($\mathcal{C}_\text{Human}$) or self-generated critique ($\mathcal{C}_\text{Self}$). $^\dagger$: models lack general question answering capability and cannot perform correction. ~*: proprietary models with unknown parameters. $^\ddagger$: measured on 265 data points rather than full test set due to annotation cost.}
    \label{tab:exp_critique}
    \vspace{-18pt}
\end{table}

\subsection{Experimental Settings}
\label{sec:exp_settings}

\begin{figure*}[htbp]
  \centering
  \begin{minipage}{0.57\textwidth}
    \centering
    \includegraphics[width=\linewidth]{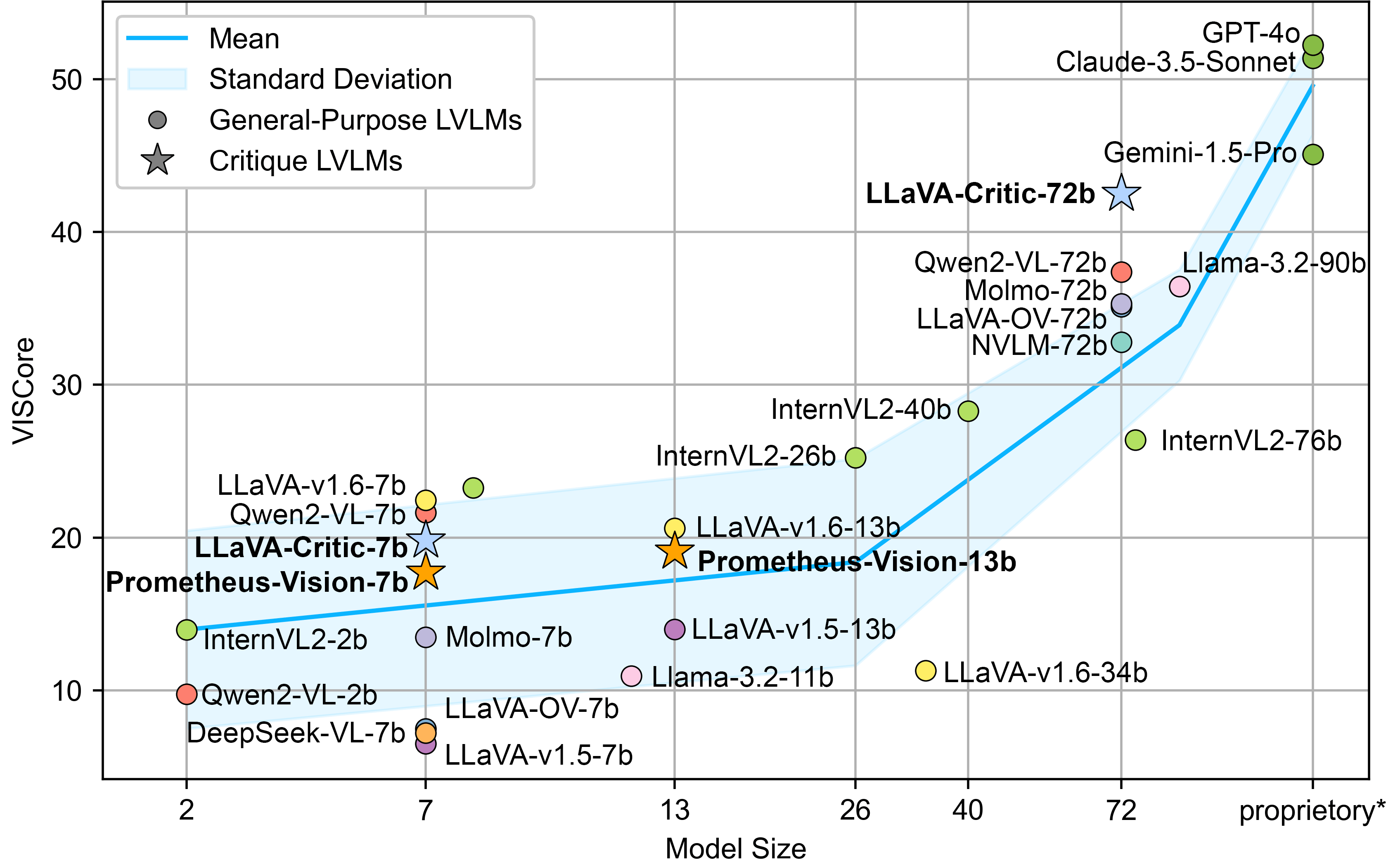}
    \vspace{-2.1em}
    \caption{\small Scaling curve for critique task.}
    \label{fig:scaling}
  \end{minipage}
  \hfill
  \begin{minipage}{0.41\textwidth}
    \centering
    \begin{minipage}{0.48\textwidth}
      \centering
      \includegraphics[width=\linewidth]{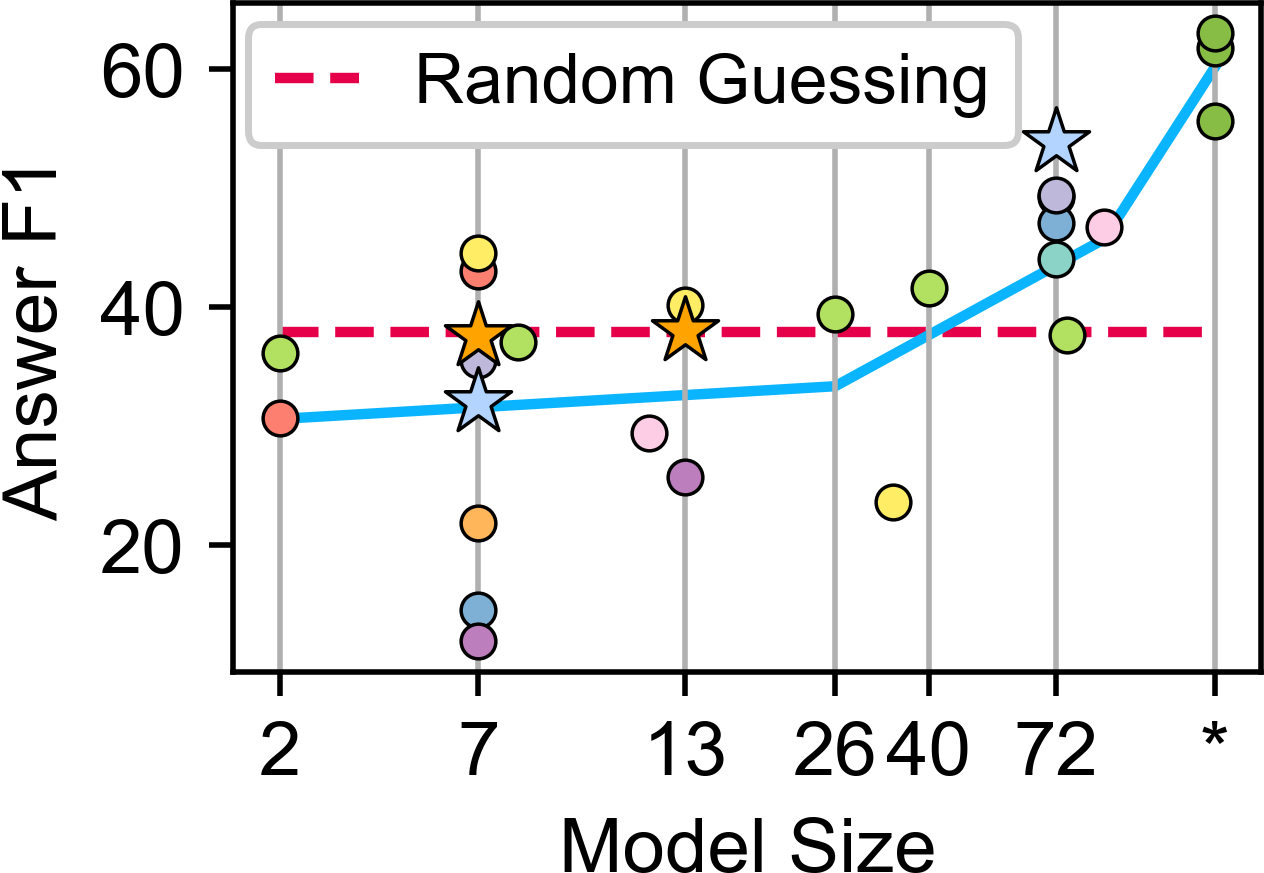}
      \vspace{-2em}
      \caption{Scaling curve for answer-level F1.}
      \label{fig:scaling_a}
    \end{minipage}
    \hfill
    \begin{minipage}{0.48\textwidth}
      \centering
      \includegraphics[width=\linewidth]{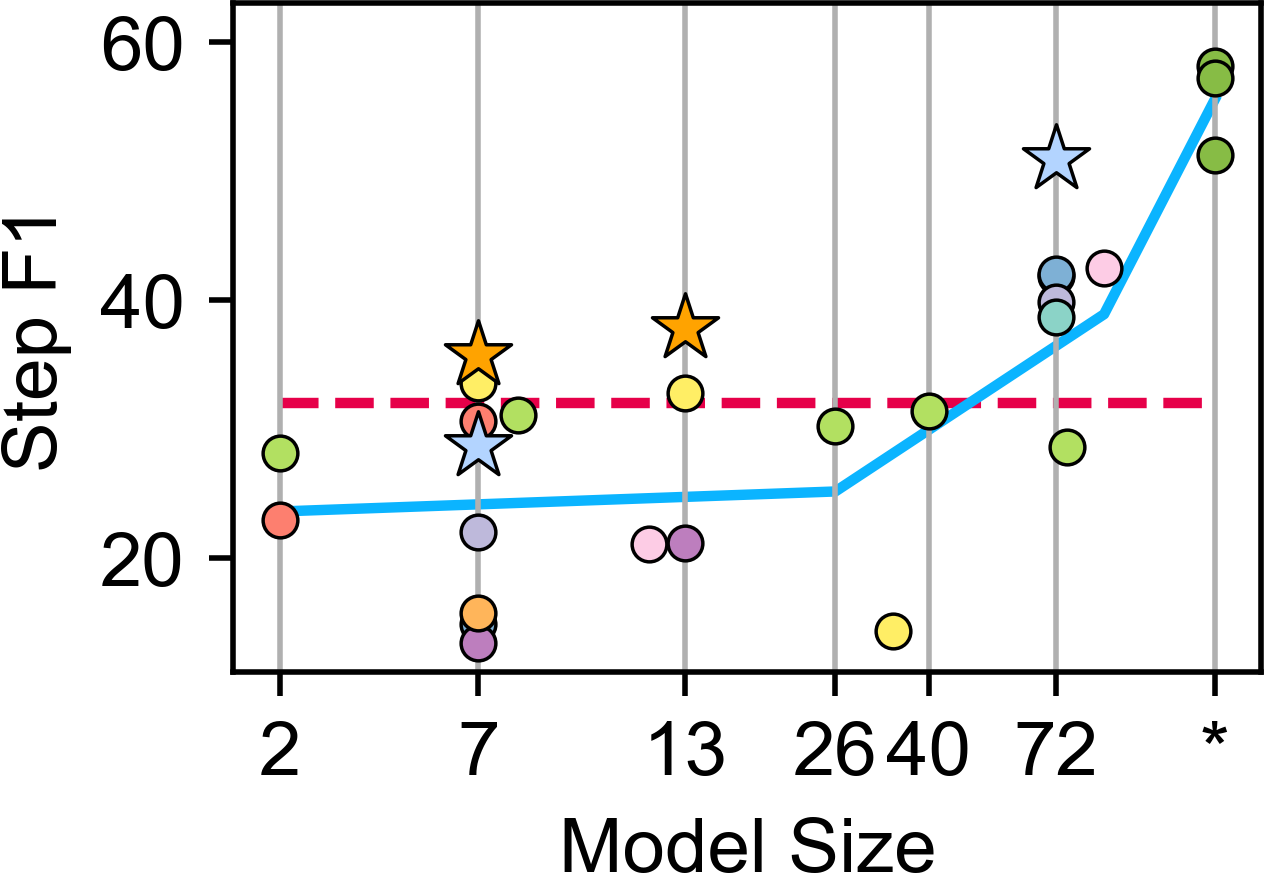}
      \vspace{-2em}
      \caption{Scaling curve for step-level F1.}
      \label{fig:scaling_t}
    \end{minipage}
    \begin{minipage}{0.48\textwidth}
      \centering
      \includegraphics[width=\linewidth]{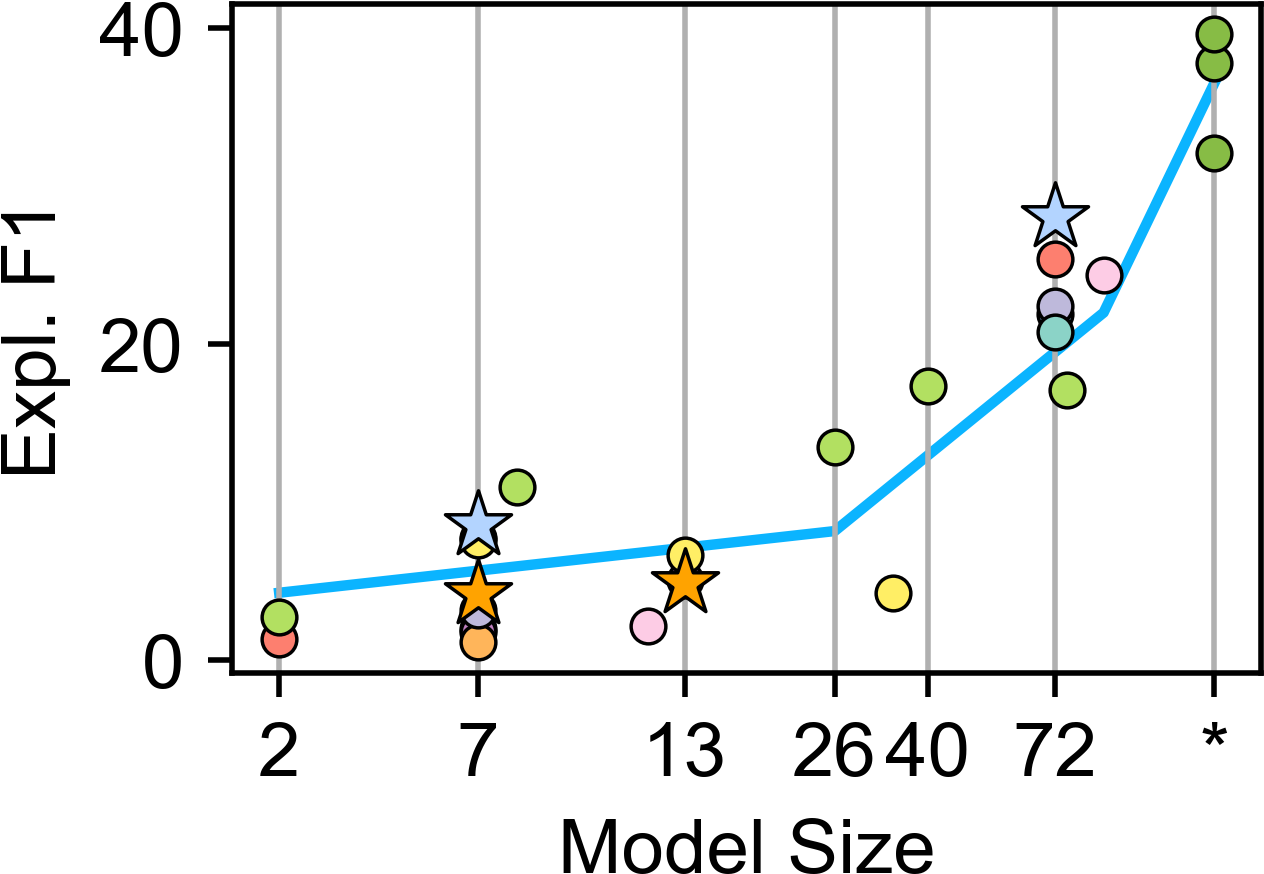}
      \vspace{-2em}
      \caption{Scaling curve for explanation-level F1.}
      \label{fig:scaling_e}
    \end{minipage}
    \hfill
    \begin{minipage}{0.48\textwidth}
      \centering
      \includegraphics[width=\linewidth]{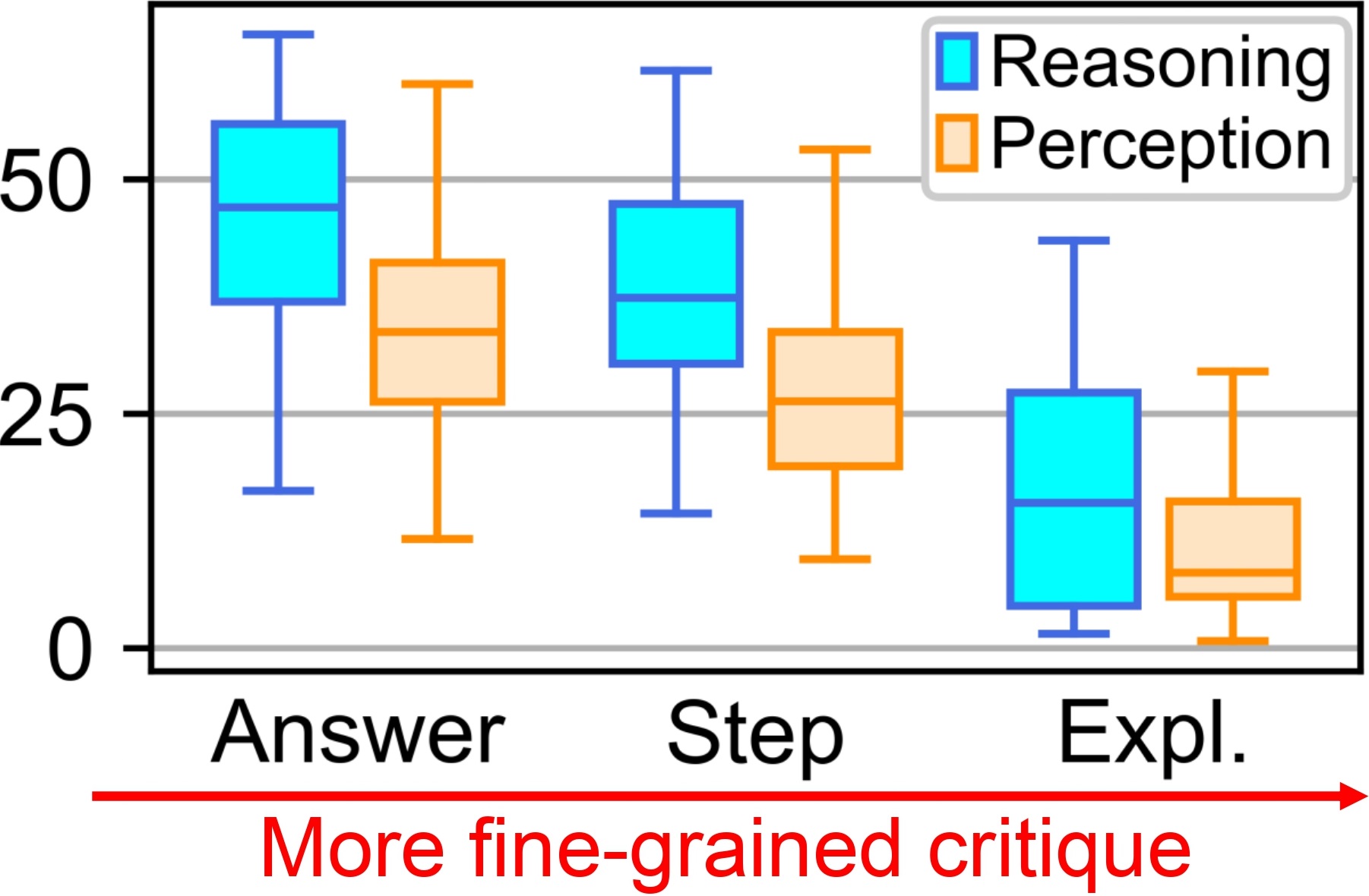}
      \vspace{-1.6em}
      \caption{Critique F1 by granularity and category.}
      \label{fig:f1_3x2}
    \end{minipage}
  \end{minipage}
    \vspace{-10pt}
\end{figure*}

\begin{figure*}[htbp]
  \centering
  \begin{minipage}{0.32\textwidth}
    \centering
    \includegraphics[width=\linewidth]{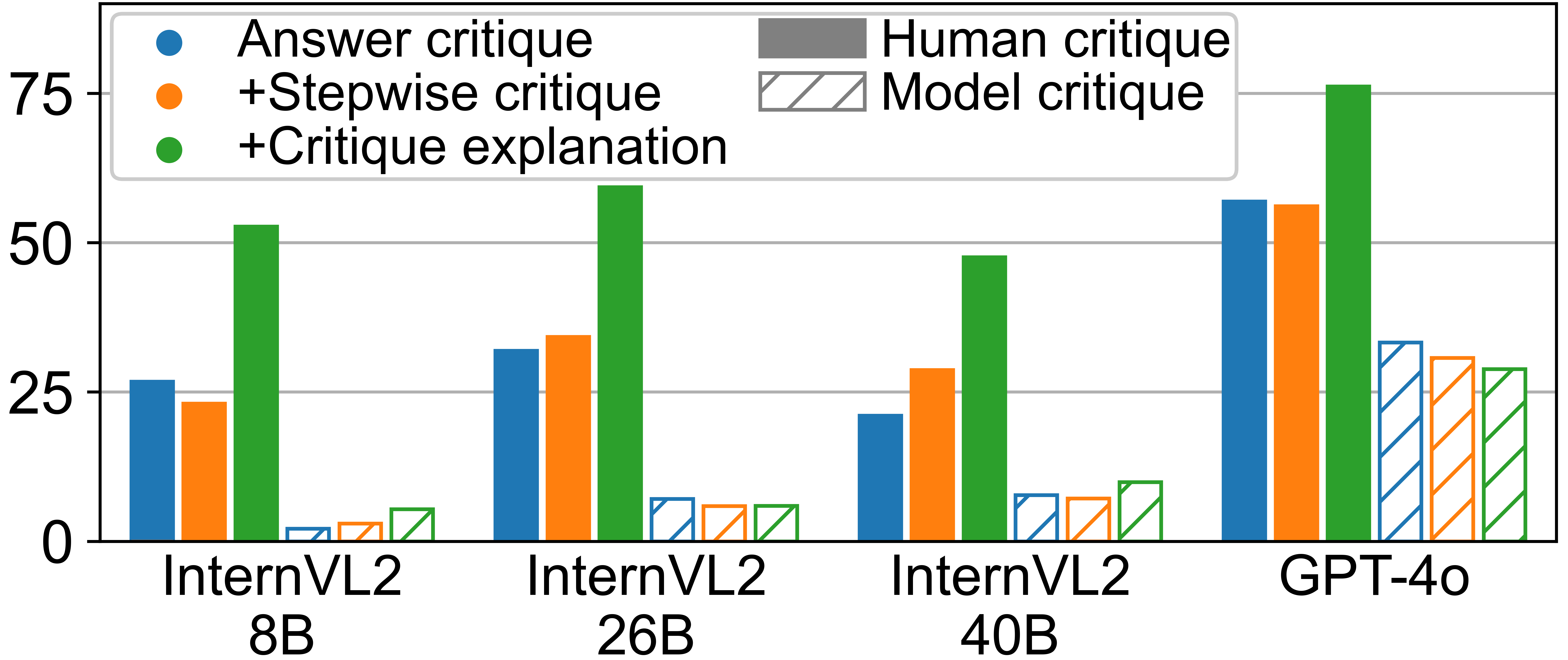}
    \vspace{-19pt}
    \caption{Correction performance given model-generated or human-generated critiques with different granularity.}
    \label{fig:correction}
  \end{minipage}
  \hfill
  \begin{minipage}{0.32\textwidth}
    \centering
    \includegraphics[width=\linewidth]{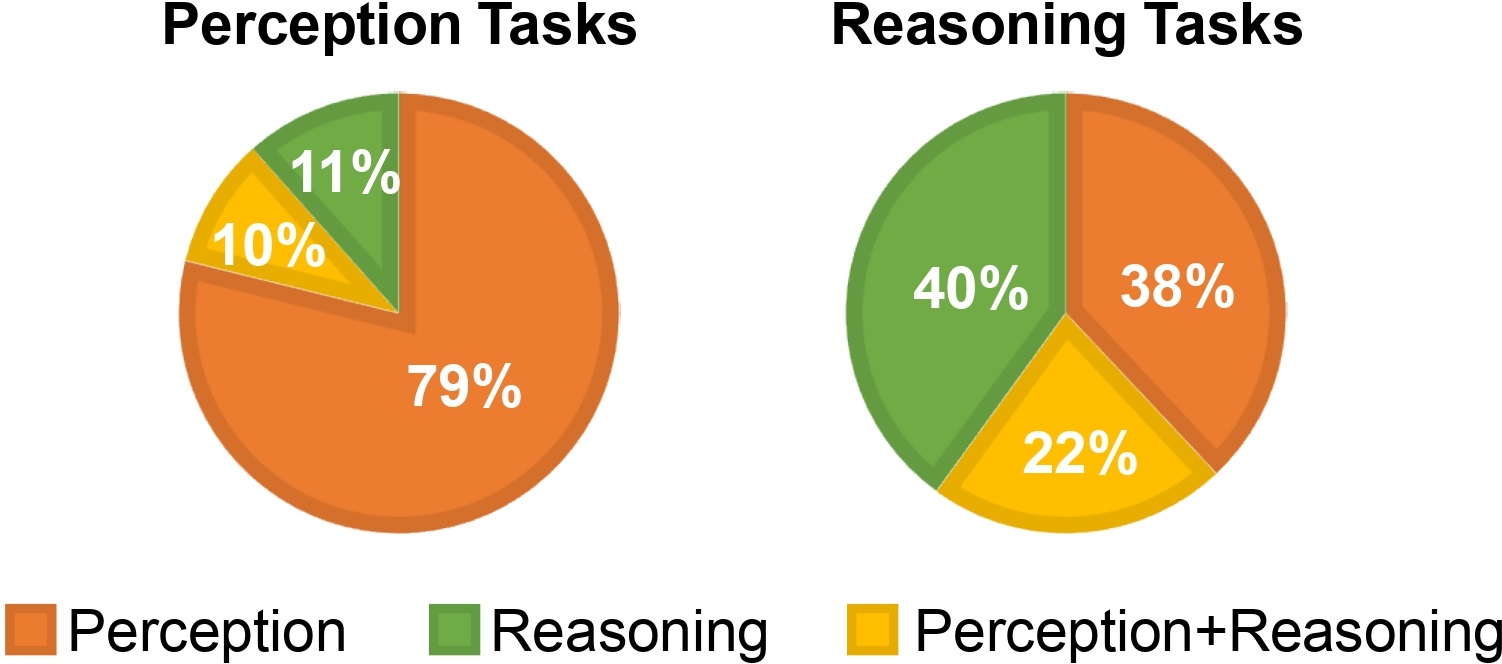}
    \vspace{-13pt}
    \caption{Error distribution in reasoning and perception tasks.}
    \label{fig:error_analysis}
  \end{minipage}
  \hfill
  \begin{minipage}{0.32\textwidth}
    \centering
    \includegraphics[width=\linewidth]{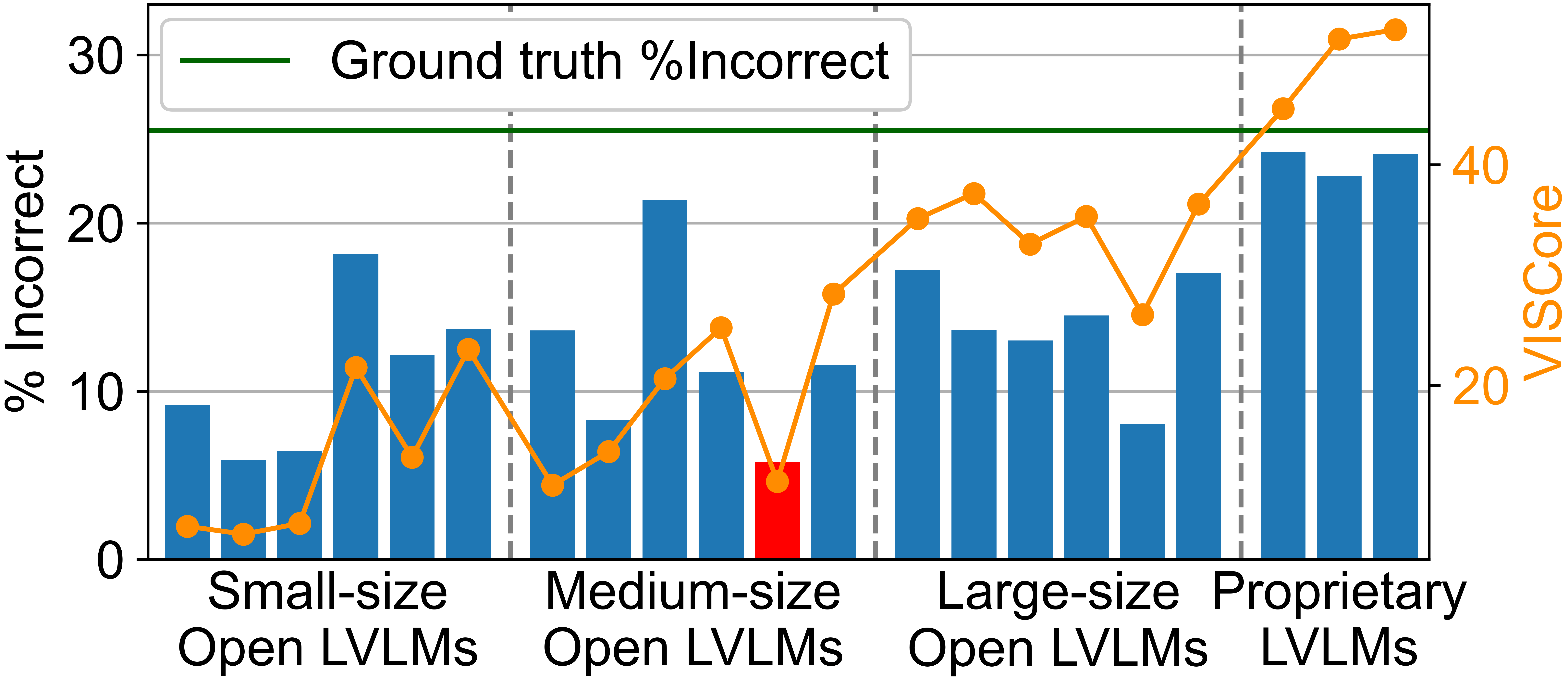}
    \vspace{-12pt}
    \caption{Percentage of data predicted as ``Incorrect''.}
    \label{fig:yesno_bias}
  \end{minipage}
  \vspace{-12pt}
\end{figure*}

We evaluate 18 open-source general-purpose LVLMs including LLaVA \citep{llava} and Llama-3.2, and three leading proprietary LVLMs including GPT-4o, Claude-3.5-Sonnet, and Gemini-Pro. We further evaluate two open LVLMs trained to provide critique, Prometheus-Vision \citep{prometheus} and LLaVA-Critic \citep{llavacritic}. As in Section \ref{sec:task_formulation}, we use \ourscore{} to evaluate critique and correction gain to evaluate correction, adopting GPT-4o for LLM-assisted evaluation. For the critique task, we further establish a human baseline performed by trained expert annotators. For the correction task, we evaluate two settings, correction based on human-annotated ground truth critique and correction based on self-generated critique to simulate the self-improvement setting. More experimental details are in Appendix \ref{sec:exp_details}.

\subsection{Results}
\label{sec:exp_main}

We report the main results in Table \ref{tab:exp_critique}. For critique capabilities, Open LVLMs significantly improve as model size scales up. Proprietary LVLMs perform better overall than open LVLMs. However, even the best proprietary LVLMs still significantly lag behind human experts. In the correction task, both top-performing open LVLMs and proprietary LVLMs can correct over 70\% errors when human-annotated critiques are available, demonstrating the substantial benefits brought by high-quality critique. In contrast, self-generated critique proves less effective and can sometimes even hurt the performance as indicated by the negative correction gains. This further verifies the importance and difficulty of producing high-quality critique for effective self-improvement.

\mypar{Non-trivial critique capability typically emerges in $\sim$70B LVLMs.} As shown in Figure \ref{fig:scaling}-\ref{fig:scaling_e}, increasing the model size consistently improves critique performance. For binary critique at answer and step levels, most of the small- and medium-sized LVLMs with $<$70B parameters perform even worse than random guessing (Figure \ref{fig:scaling_a}, \ref{fig:scaling_t}). The correlation between model size and performance is especially strong for explanation-level F1 (Figure \ref{fig:scaling_e}), as smaller models may correctly predict binary critiques but fail to provide coherent explanations. Overall, non-trivial (above-random) critique capabilities typically emerge in 70B-level or proprietary LVLMs.

\mypar{Benefits and challenges of and fine-grained critique.} To investigate whether our proposed fine-grained critique contributes to self-improvement, we show the correction gain with critiques of different granularity in Figure \ref{fig:correction} and \ref{fig:correction_full}. %
When using human-annotated high-quality critique, fine-grained critique with step-wise labels and explanations consistently boosts the performance. However, this trend does not hold for model-generated critique, where fine-grained critique does not consistently lead to improvements and sometimes even hurts the performance. As shown in Figure \ref{fig:f1_3x2}, the critique performance of LVLMs decreases as the required level of granularity increases. This highlights that the effective generation of fine-grained critique is both crucial and challenging.

\begin{figure*}[htbp]
  \centering
  
  \begin{minipage}{0.32\textwidth}
    \centering
    \includegraphics[width=\linewidth]{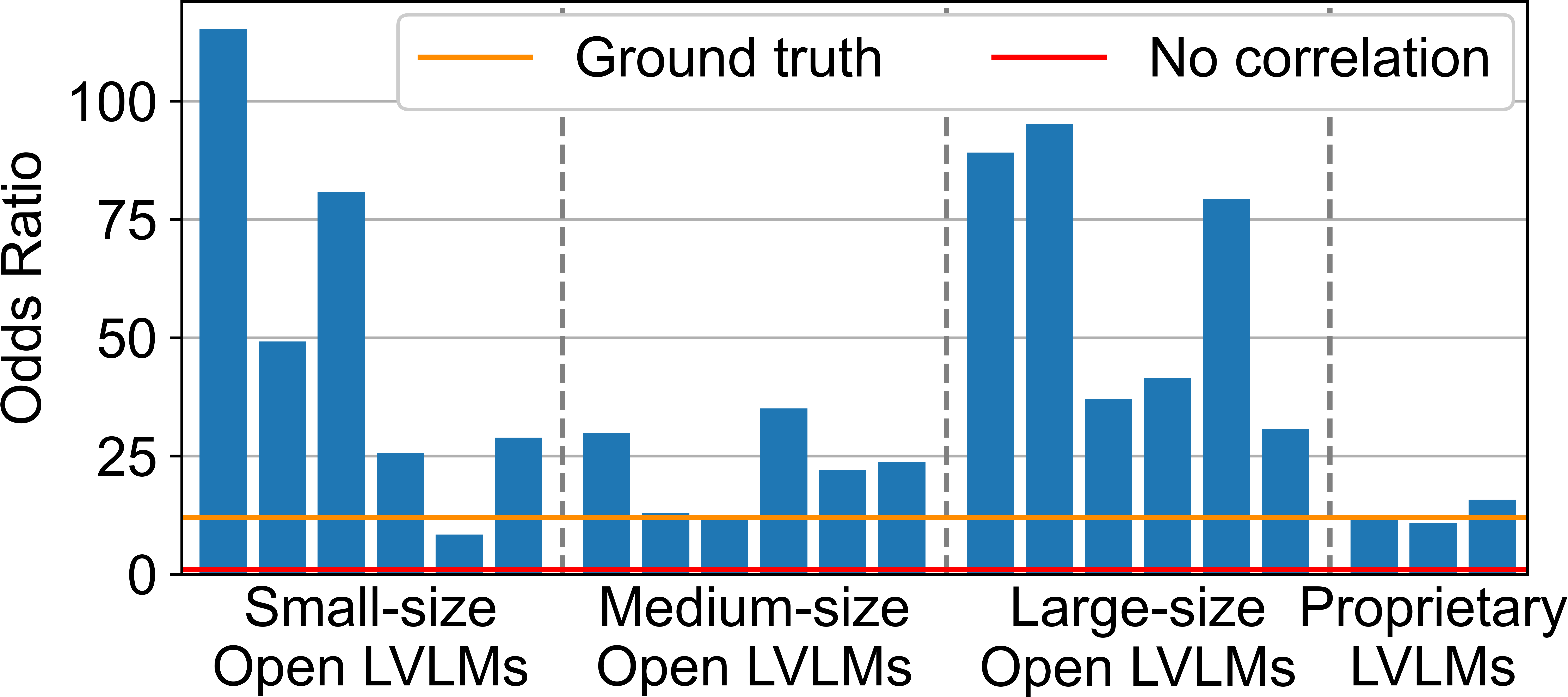}
    \vspace{-1.75em}
    \caption{Odds ratio measuring the likelihood of a future step being incorrect given that a prior step is incorrect (higher values indicates a greater likelihood).}
    \label{fig:or}
  \end{minipage}
  \hfill
  \begin{minipage}{0.32\textwidth}
    \centering
    \includegraphics[width=\linewidth]{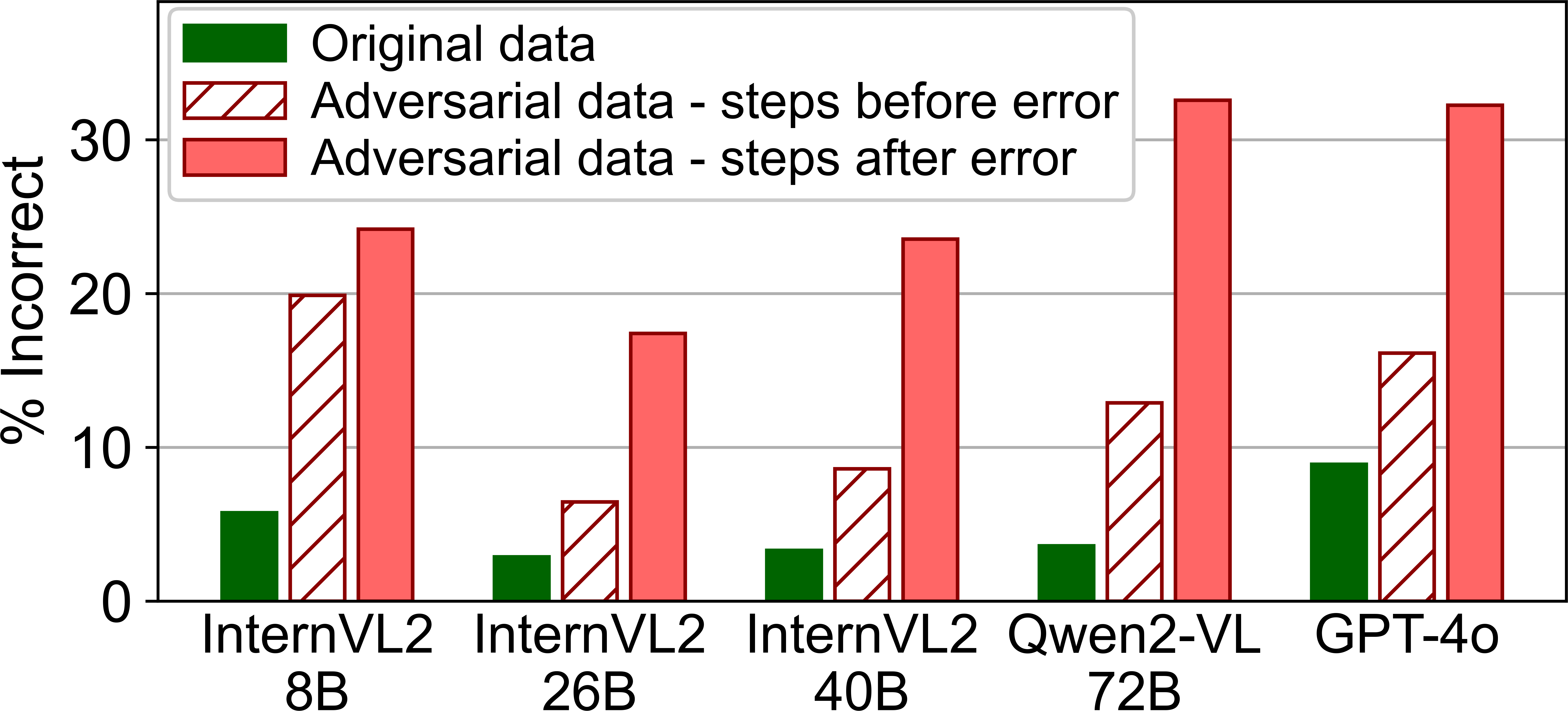}
    \vspace{-1.9em}
    \caption{Percentage of data predicted as ``Incorrect'' on a subset that is all correct. We show the statistics on original data and adversarially constructed data.}
    \label{fig:adversarial}
  \end{minipage}
  \hfill
  \begin{minipage}{0.32\textwidth}
    \centering
    \vspace{-1.15em}
    \includegraphics[width=\linewidth]{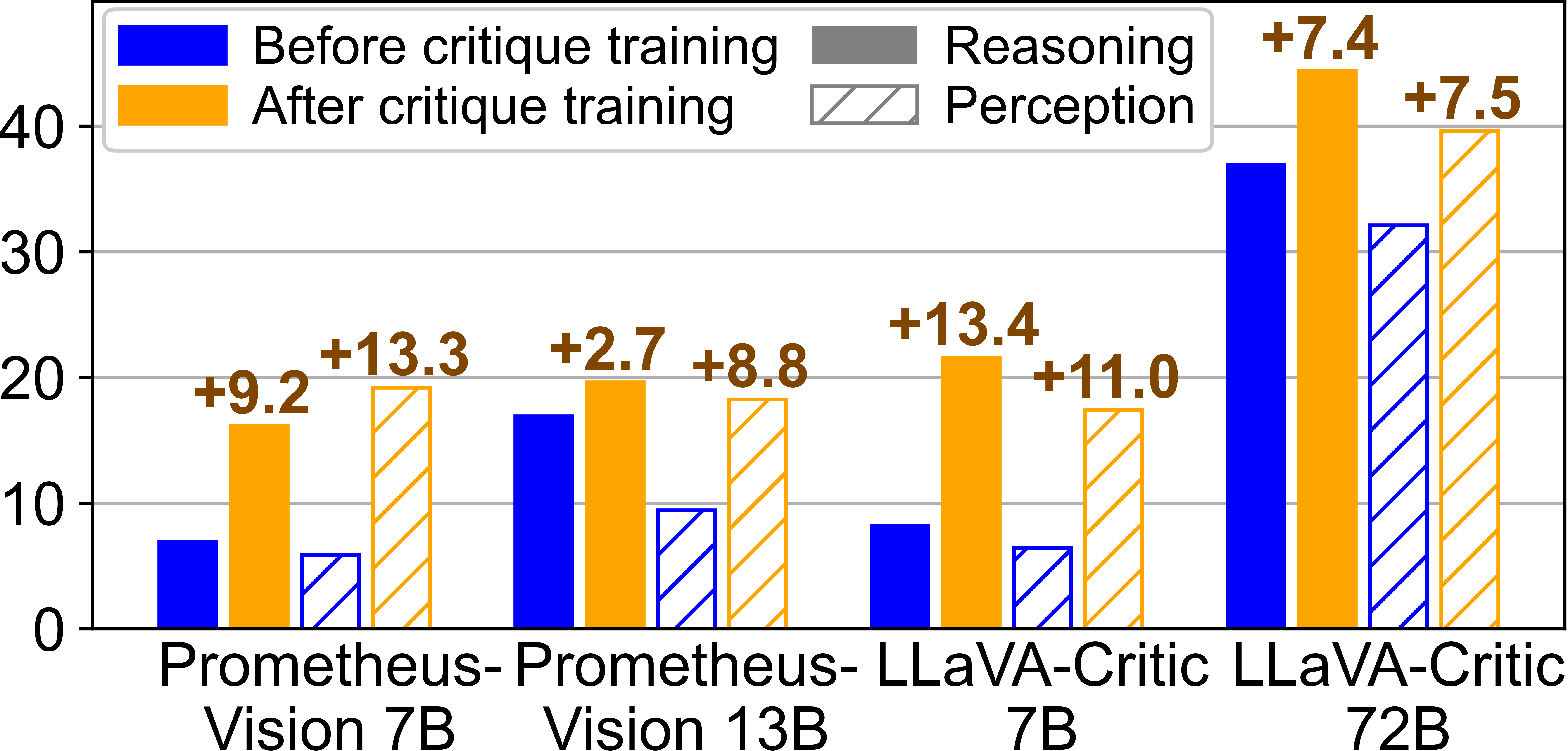}
    \caption{Critique performance before and after critique training.}
    \label{fig:critic_barplot}
  \end{minipage}
  \vspace{-18pt}
\end{figure*}

\begin{figure}[h]
    \centering
    \includegraphics[width=\linewidth]{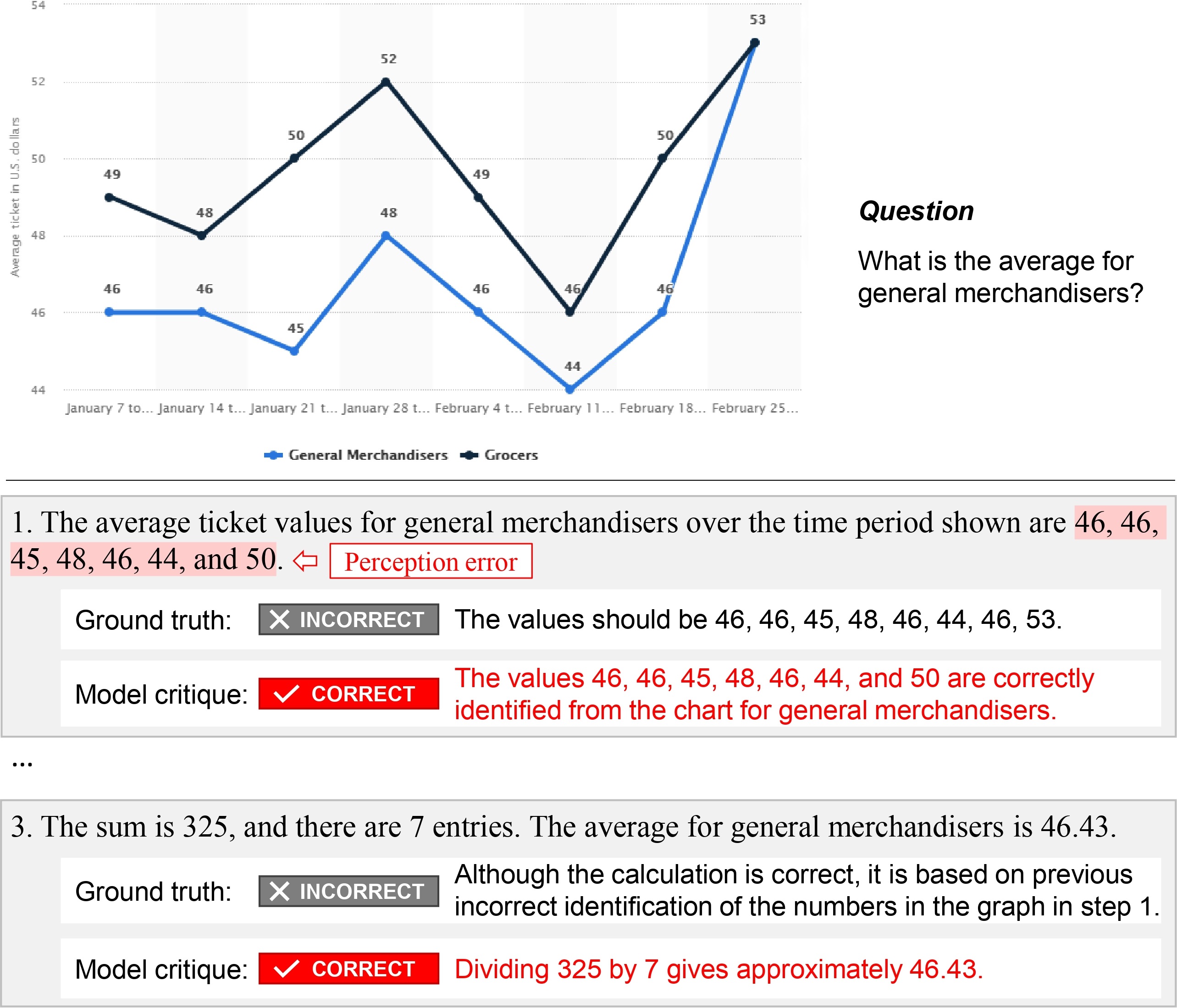}
    \vspace{-2em}
    \caption{\textbf{Example of failure to critique visual perception.} Step 1 makes a perception error when identifying the values, but the model critique fails to identify the error.}
    \label{fig:case1}
    \vspace{-18pt}
\end{figure}

\mypar{Patterns of critique failure.} We further analyze the patterns when LVLMs fail to provide accurate critique and summarize three common patterns as follows:%

\underline{Failure to critique visual perception.} As in Figure \ref{fig:f1_3x2}, critique performance on perception tasks is consistently lower than on reasoning tasks at every level, highlighting a unique challenge in visual reasoning to critique visual perception rather than reasoning. Notably, this trend is different from human performance, as human experts perform slightly better on perception tasks (\ourscore{} = 88.5 for perception, 84.6 for reasoning). To further investigate this issue, we manually review 100 examples and categorize the sources of critique errors into three types: perception errors, reasoning errors, and a combination of both reasoning and perception errors. The results are shown in Figure \ref{fig:error_analysis}. Unsurprisingly, the majority of errors in perception tasks are due to perception errors. Interestingly, we find that a significant portion of errors in reasoning tasks are also caused by perception (38\%) or a combination of perception and reasoning (22\%). This indicates that visual perception remains a bottleneck even for reasoning-intensive problems. An example is shown in Figure \ref{fig:case1}, where the model fails to critique the incorrect digit recognition in step 2, resulting in an incorrect critique.

\begin{figure}[h]
        \centering
        \includegraphics[width=\linewidth]{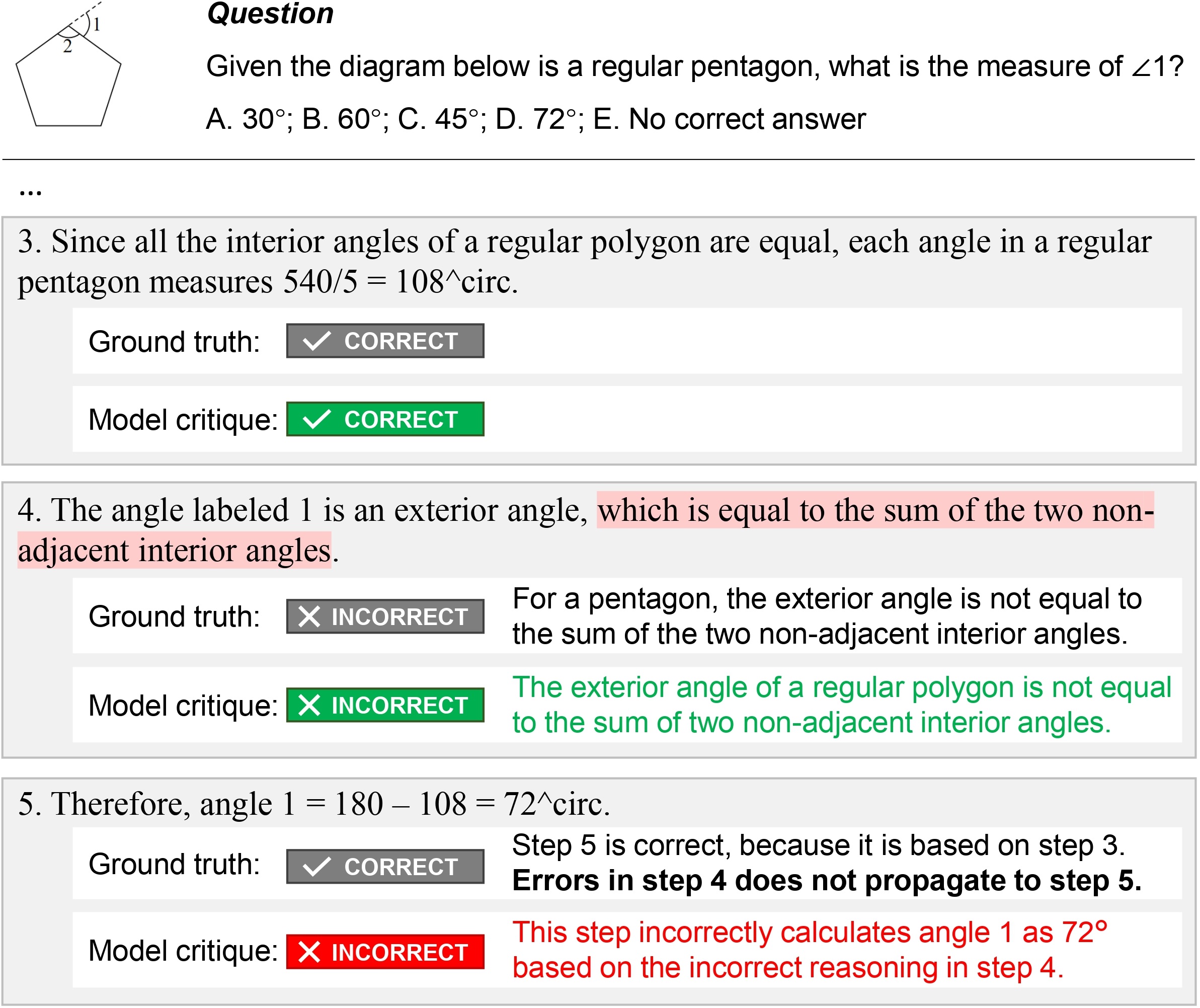}
        \vspace{-1.7em}
        \caption{\textbf{Example of exaggerated assumption of error propagation.} The model critique mistakenly believes the error in step 4 is propagated to step 5, while in fact step 5 is based on step 3 and thus independent from the error in step 4.}
        \label{fig:case2}
    \vspace{-16pt}
\end{figure}

\underline{Reluctance to ``say no''.} We observe that LVLMs generally show a strong bias against ``saying no'', meaning they are more likely to judge an answer or step as correct rather than incorrect. As shown in Figure \ref{fig:yesno_bias}, all LVLMs predict ``Incorrect'' less frequently than the ground truth annotations. Some models like LLaVA-1.6-34B identify an extremely low percentage of inputs as incorrect (5.7\% for LLaVA-1.6-34B), resulting in notably poor performance. This observation aligns with existing work which similarly reports a tendency for LVLMs to favor ``Yes" over ``No" in binary classification tasks \citep{pope,hallusionbench}. As pointed out by \citep{liumitigating,hallusionbench}, a potential reason for this bias is imbalanced instruction tuning data, as instruction tuning typically encourages models to agree with users.

\underline{Exaggerated assumption of error propagation.} In the sequential chain-of-thought, errors may naturally propagate from earlier steps to later ones. However, we observe that LVLMs exhibit a much stronger bias in expecting errors to propagate. We use the odds ratio to measure the association between errors in earlier steps and those in later steps, and thus measuring the error propagation effect. As shown in Figure \ref{fig:or}, the step-wise correctness judged by LVLMs indicates a much stronger error propagation effect than the ground truth data. This causes LVLMs to struggle when later steps are independent of earlier mistakes; an example of this is shown in Figure \ref{fig:case2}. To further quantitatively assess the effect on critique performance, we construct an adversarial test set by prompting LLMs to inject errors into the middle sentence of an otherwise correct CoT. As shown in Figure \ref{fig:adversarial}, LVLMs fail to recognize that the error is independent of its context, thus mistakenly identify significantly more steps as incorrect, especially for the steps after the injected error.

\mypar{Critique training helps.} In addition to general-purpose LVLMs, we evaluate two models specifically fine-tuned to provide critique: Prometheus-Vision \citep{prometheus} and LLaVA-Critic \citep{llavacritic}. As shown in Figure \ref{fig:scaling}, the critique LVLMs (\includegraphics[height=\fontcharht\font`\B]{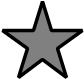}) consistently perform above average relative to their corresponding sizes. Notably, the 72B LLaVA-Critic model significantly outperforms other 70B-level LVLMs and ranks as the best across all open LVLMs.
The benefits are more substantial when comparing the critique LVLMs to the base models they were fine-tuned from. %
As in Figure \ref{fig:critic_barplot}, critique training significantly improves the performance on \ours{}, highlighting specialized training as a promising strategy to boost self-improvement capabilities. When applied to correction, the critique generated by LLaVA-Critic-72B also brings the best correction performance across all open LVLMs.

\begin{figure}
    \centering
    \includegraphics[width=0.95\linewidth]{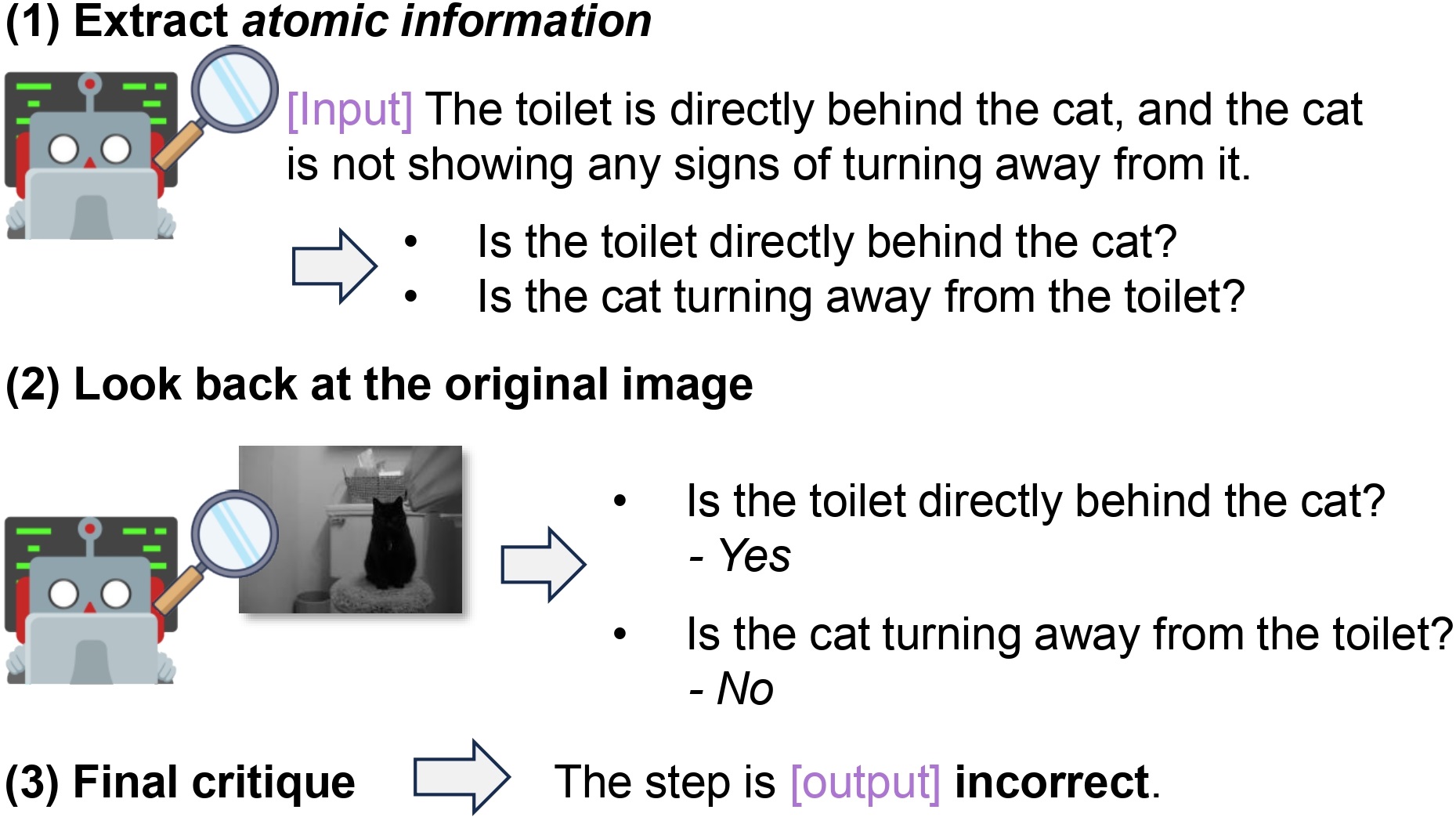}
    \vspace{-0.6em}
    \caption{\textbf{Illustration of \ourscritic{} method.}}
    \label{fig:ours_critic}
    \vspace{-16pt}
\end{figure}

\section{\ourscritic{} as Improved Baseline}
\label{sec:ours_method}

In this section, we present an initial step toward enhancing the critique capabilities of LVLMs.
As in Section \ref{sec:exp_main}, the failure to critique visual perception and the reluctance to ``say no'' are two major bottlenecks for effective critique. To prevent LVLMs from blindly agreeing with the visual perception in the model response, we propose to explicitly check the image and verify the visual information before performing the critique. 

An illustration of our proposed \ourscritic{} is shown in Figure \ref{fig:ours_critic}, with the formal algorithm in Alg. \ref{alg:ours}. \ourscritic{} iteratively analyzes each step in the CoT, extracting \textbf{atomic information} and converting it into question format (L3). To ensure thoroughness, it incorporates the original question $q$ into the question set $\mathcal{Q}$ (L6). The extracted information is then verified by prompting these questions against the original image $I$ (L9), and the generated answers are synthesized into a holistic critique (L12). By breaking down complex reasoning into verifiable units and iteratively revisiting the image for validation, \ourscritic{} better mirrors the human critique process, where examiners cross-reference individual claims with visual evidence.

\begin{table}[!t]
    \centering
    \small
\setlength{\tabcolsep}{3.5pt}
    \begin{tabular}{l|rrrr|r}
    \toprule
& \multicolumn{4}{c|}{Critique} & Correction \\
& Total & {Reas.} & {Perc.} &\%Inc. & with $\mathcal{C}_\text{self}$ \\
\midrule
InternVL2-8B & 23.3 & 28.2 & 14.7 & 13.7 & 5.5 \\
~~~~+ \ourscritic{} & \textbf{30.7} & \textbf{31.5} & \textbf{29.5} & 32.0 & \textbf{11.7}\vspace{-4pt} \\
\scriptsize\color[RGB]{0,120,0}& \scriptsize\color[RGB]{0,120,0} (+7.4) &\scriptsize\color[RGB]{0,120,0} (+3.2) & \scriptsize\color[RGB]{0,120,0}(+14.8) & & \scriptsize\color[RGB]{0,120,0}(+6.2) \\
\midrule
InternVL2-26B & 25.2 & 31.2 & 15.0 & 11.1 & 6.0 \\
~~~~+ \ourscritic{} & \textbf{38.7} & \textbf{39.6} & \textbf{37.2} & 31.4 & \textbf{14.4}\vspace{-4pt} \\
\scriptsize\color[RGB]{0,120,0}& \scriptsize\color[RGB]{0,120,0} (+13.5) &\scriptsize\color[RGB]{0,120,0} (+8.4) & \scriptsize\color[RGB]{0,120,0}(+22.2) & & \scriptsize\color[RGB]{0,120,0}(+8.4) \\
\midrule
Gemini-1.5-Pro & 45.0 & 48.7 & 39.3 & 24.2 & 24.9 \\
~~~~+ \ourscritic{} & \textbf{55.5} & \textbf{61.1} & \textbf{47.4} & 33.6 &  \textbf{36.4}\vspace{-4pt} \\
\scriptsize\color[RGB]{0,120,0}& \scriptsize\color[RGB]{0,120,0} (+10.5) &\scriptsize\color[RGB]{0,120,0} (+12.4) & \scriptsize\color[RGB]{0,120,0}(+8.1) & &  \scriptsize\color[RGB]{0,120,0}(+11.5) \\
\midrule
GPT-4o & 52.2 & 55.7 & 46.8 & 24.1 & 28.8 \\
~~~~+ \ourscritic{} & \textbf{57.7} & \textbf{60.0} & \textbf{54.2} & 33.2 &  \textbf{34.9}\vspace{-4pt} \\
\scriptsize\color[RGB]{0,120,0}& \scriptsize\color[RGB]{0,120,0} (+5.5) &\scriptsize\color[RGB]{0,120,0} (+4.3) & \scriptsize\color[RGB]{0,120,0}(+8.6) & &  \scriptsize\color[RGB]{0,120,0}(+6.1) \\
\bottomrule
    \end{tabular}
    \vspace{-.9em}
    \caption{\textbf{Performance of \ourscritic{} on critique and correction tasks.} Critique is evaluated by \ourscore{}, and correction is evaluated by correction ratio. We also report the \ourscore{} for reasoning (Reas.) and perception (Perc.) tasks, and the percentage of inputs predicted as incorrect (\%Inc.).}
    \label{tab:exp_correction_ours}
    \vspace{-16pt}
\end{table}

As shown in Table \ref{tab:exp_correction_ours}, our proposed \ourscritic{} significantly enhances the critique performance for four leading LVLMs. The gains are more significant for perception tasks (+13.4 on average for perception v.s. +7.1 for reasoning), showing that \ourscritic{} effectively addresses the challenges in critiquing visual perception. Additionally, we find that \ourscritic{} identifies a higher percentage of inputs as incorrect, thereby alleviating the reluctance to disagree. When applied to the correction task, critique generated by \ourscritic{} brings notable improvements.

\section{Conclusions}

This paper presents \ours{}, a novel benchmark designed to evaluate LVLMs in terms of critique and correction, two fundamental capabilities for self-improvement. \ours{} evaluates fine-grained critique that identifies the correctness of each reasoning step in a chain-of-thought and further provides natural language explanations. We curate the dataset from a diverse range of 18 datasets across 8 tasks and extensively evaluate 24 LVLMs. Our experiments show while LVLMs perform well in correction given high-quality critique, the effective generation of critique is the crucial bottleneck for self-improvement. We further investigate the model performance on critique tasks and identify three main issues contributing to LVLMs' critique failures: failure to critique visual perception, reluctance to ``say no'', and the exaggerated assumption of error propagation. To address these limitations, we designed the \ourscritic{} strategy, which can be combined with leading LVLMs to enhance the critique capability by up to 13.5 and further benefit the correction performance by up to 11.5. This work serves as an initial exploration into self-improvement of visual reasoning, aiming to inspire future research in this direction.

\section*{Acknowledgment}

This research is based upon work supported by CISCO, U.S. DARPA ECOLE Program No. \#HR00112390060, and OFFICE OF NAVAL RESEARCH Award \#N00014-23-1-2780. The views and conclusions contained herein are those of the authors and should not be interpreted as necessarily representing the official policies, either expressed or implied, of DARPA, or the U.S. Government. The U.S. Government is authorized to reproduce and distribute reprints for governmental purposes notwithstanding any copyright annotation therein.

{
    \small
    \bibliographystyle{ieeenat_fullname}
    \bibliography{main}
}

\appendix
\clearpage
\setcounter{page}{1}
\twocolumn[\centering\Large\textbf{Appendix}\vspace{1.5em}]

\section{Dataset Examples}
\label{sec:dataset_examples}

\begin{figure*}
    \centering
\includegraphics[width=\linewidth]{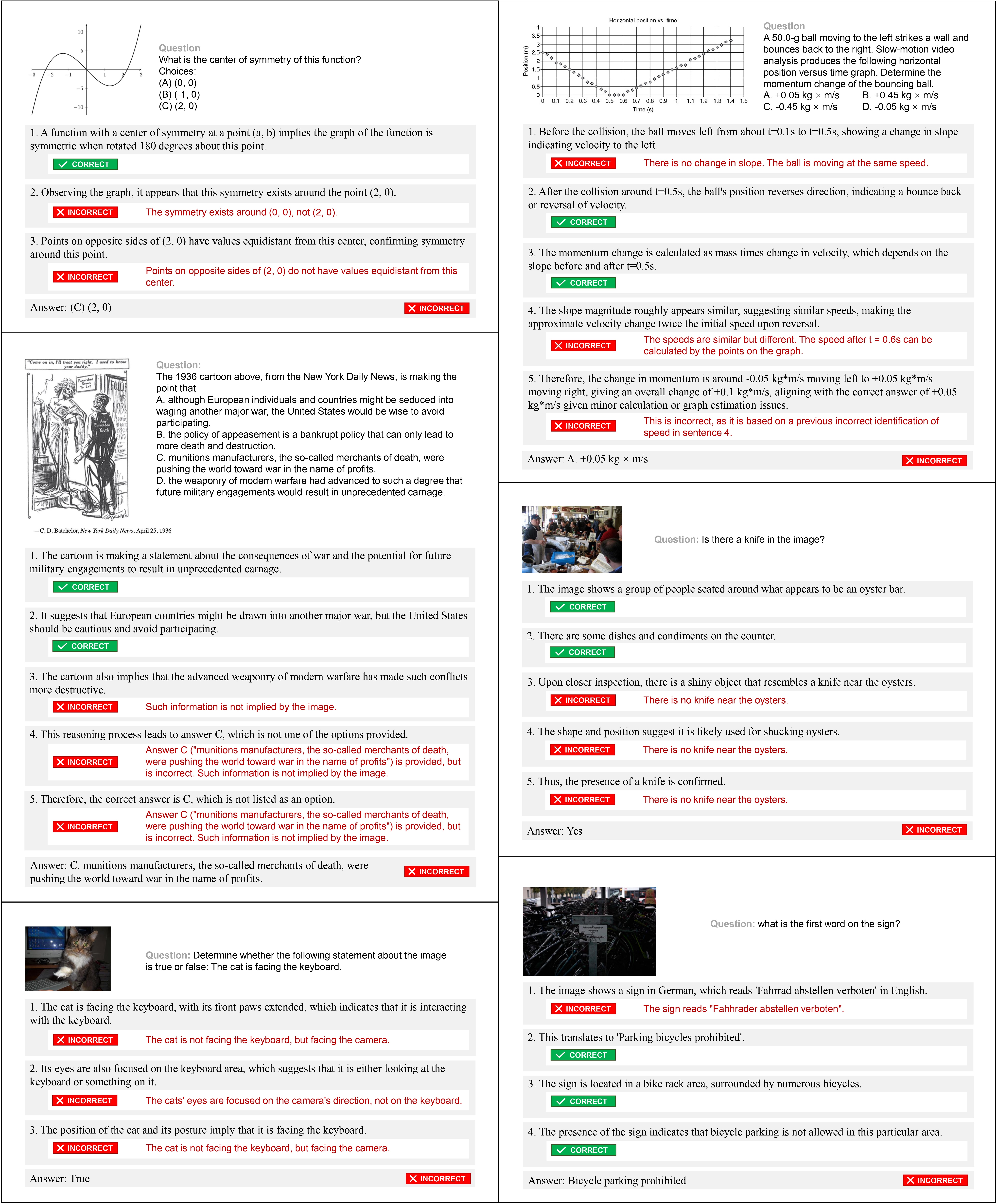}
    \caption{Dataset examples.}
    \label{fig:examples}
\end{figure*}

\begin{figure*}
    \centering
\includegraphics[width=\linewidth]{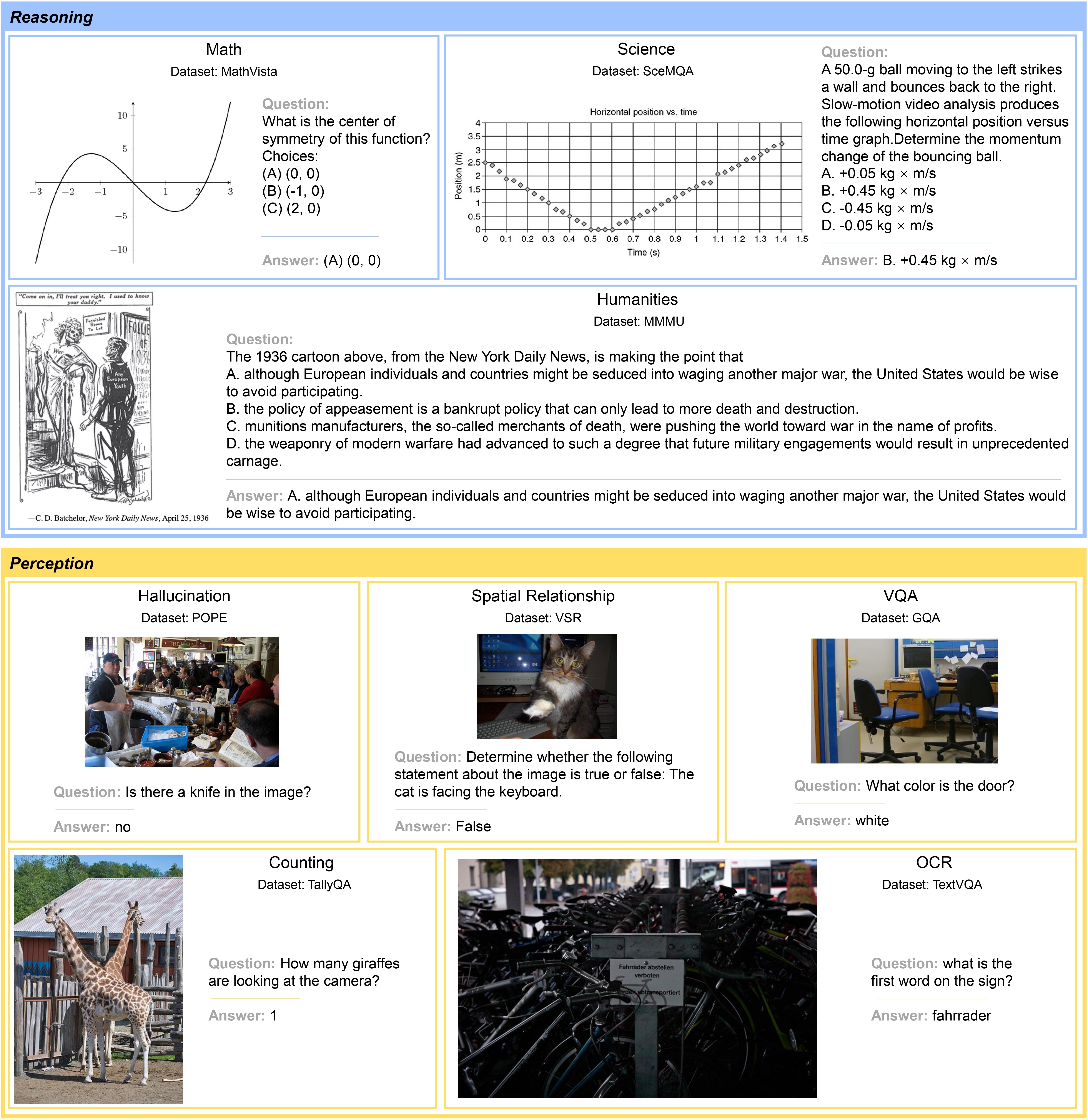}
    \caption{Examples for each task in our dataset.}
    \label{fig:task_input}
\end{figure*}

We present examples of \ours{} dataset in Figure \ref{fig:examples}.
We further show the details of each step in the dataset construction process as follows:

\mypar{Task input collection.} We collect images, questions and ground truth answers from existing visual question answering datasets. The distribution of our tasks and datasets are shown in Figure 3 in the main content. Examples from each task are in Figure \ref{fig:task_input}.

\mypar{Response collection.} We use 7 LVLMs to sample model responses, including both the CoT and the final answer. We prompt the LVLMs to generate CoT with less than five sentences, and additionally remove CoT with more than five sentences. The prompt template and an example output are shown in Figure \ref{fig:response_collection}.

\begin{figure*}
    \centering
\includegraphics[width=\linewidth]{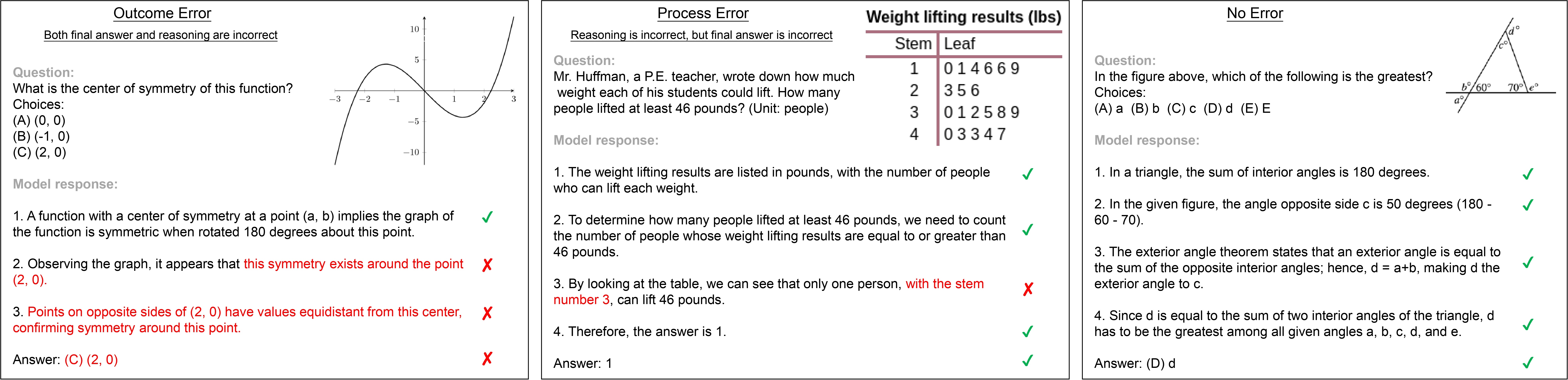}
    \caption{Examples of model responses with outcome errors, process errors, and no errors.}
    \label{fig:response_filtering}
\end{figure*}

\mypar{Response filtering.} In the response filtering stage, we filter the responses into three subsets: responses with outcome errors, responses with process errors, and responses with no errors. Examples from each of the three subsets are shown in Figure \ref{fig:response_filtering}.

\mypar{Critique collection.} We train three human annotators to provide dense and fine-grained critique with a binary label for each step, and a natural language explanation if the step is considered as incorrect. The annotation interface is shown in Figure \ref{fig:critique_collection_ui}.

\section{Metric Design Details}
\label{sec:metric}

\mypar{\ourscore{} design.}
To verify the robustness of \ourscore{} metric design, we conducted human evaluation on 200 data where annotators compare pairs of model generated critique and judge which is better.
We find that \ourscore{} correlates the best with human judgment ($\tau=0.58$), outperforming alternatives like arithmetic mean of F1 ($\tau=0.42$) and exlanation BLEU ($\tau=0.41$). This is because \ourscore{} prioritizes explanation correctness, better aligning with human focus, and LLM-based evaluation of explanation is more effective than BLEU.

\mypar{LLM evaluation.} The calculation of explanation-level F1 requires matching model-generated critique explanations against human-annotated explanations using LLMs. We use GPT-4o for this evaluation, and the prompt is shown in Figure \ref{fig:prompt_gpt_eval}.
To evaluate the reliability of this evaluation, we manually compare 100 model-generated explanations against human explanations, and evaluate the agreement between our manual evaluation and LLM evaluation. We observe a high agreement of 0.80 accuracy and 0.61 Cohen's Kappa, validating the robustness of LLM evaluation.

\mypar{Metric calculation examples.} To better illustrate the metric calculation, we present two examples in Figure \ref{fig:viscore} and \ref{fig:cg}, illustrating the calculation of \ourscore{} for critique task and correction gain for correction task respectively.

\section{Experimental Details}
\label{sec:exp_details}

We evaluate 27 open LVLMs and 3 proprietary LVLMs. For the open LVLMs, we evaluate models from 7 families: DeepSeek-VL \citep{deepseekvl}, LLaVA \citep{llava}, InternVL \citep{internvl}, Qwen-VL \citep{qwen2vl}, Molmo \citep{molmo}, NVLM \citep{nvlm}, Llama-3.2, and MiniCPM \citep{yao2024minicpm}. Specifically, we run the inference with fast serving frameworks. Specifically, we evaluate Qwen2-VL, Molmo, Llama-3.2, NVLM and MiniCPM with \texttt{vllm}, evaluate LLaVA-OV and LLaVA-Critic with \texttt{sglang}, and evaluate InternVL2, DeepSeek-VL, LLaVA-v1.5, LLaVA-v1.6, Qwen-VL and Prometheus-Vision with \texttt{lmdeploy}. We set the sampling temperature as 0.7. %
For \ourscritic{}, we use the same set of hyperparameters. The prompt for critique and correction are in Figure \ref{fig:prompt_critique} and \ref{fig:prompt_correction}.

For the human baseline, we ask one of the trained annotators to establish the human baseline on a randomly selected subset of 265 data points. To reduce annotation costs, we provide the ground truth answers to the annotator. The 100\% answer-level F1 shows that the ground truth answers are verified by the annotator to be correct.

\section{Additional Results}

\begin{figure*}
    \centering
\includegraphics[width=\linewidth]{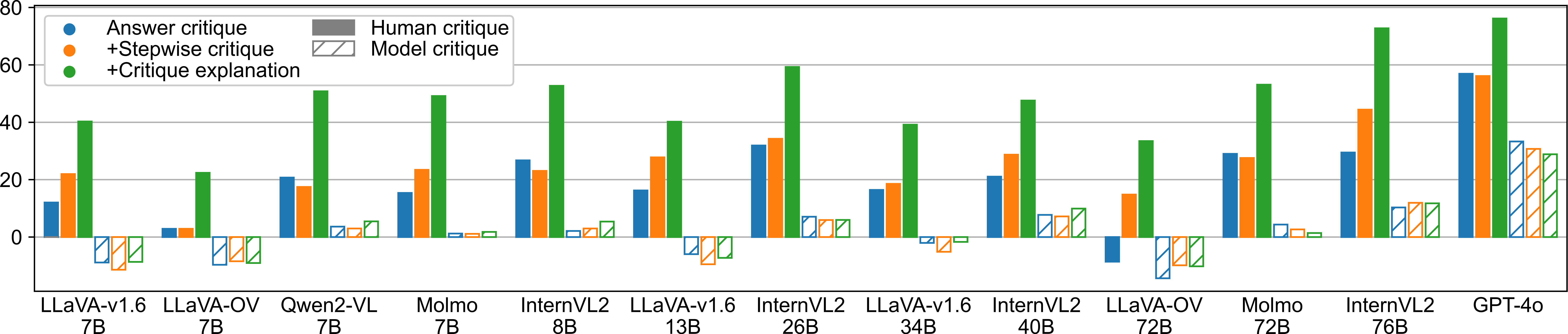}
    \caption{Correction performance given model-generated or human-generated critiques with different granularity.}
    \label{fig:correction_full}
\end{figure*}

\begin{figure*}
    \centering
    \begin{minipage}{0.48\linewidth} %
        \centering
\includegraphics[width=\linewidth]{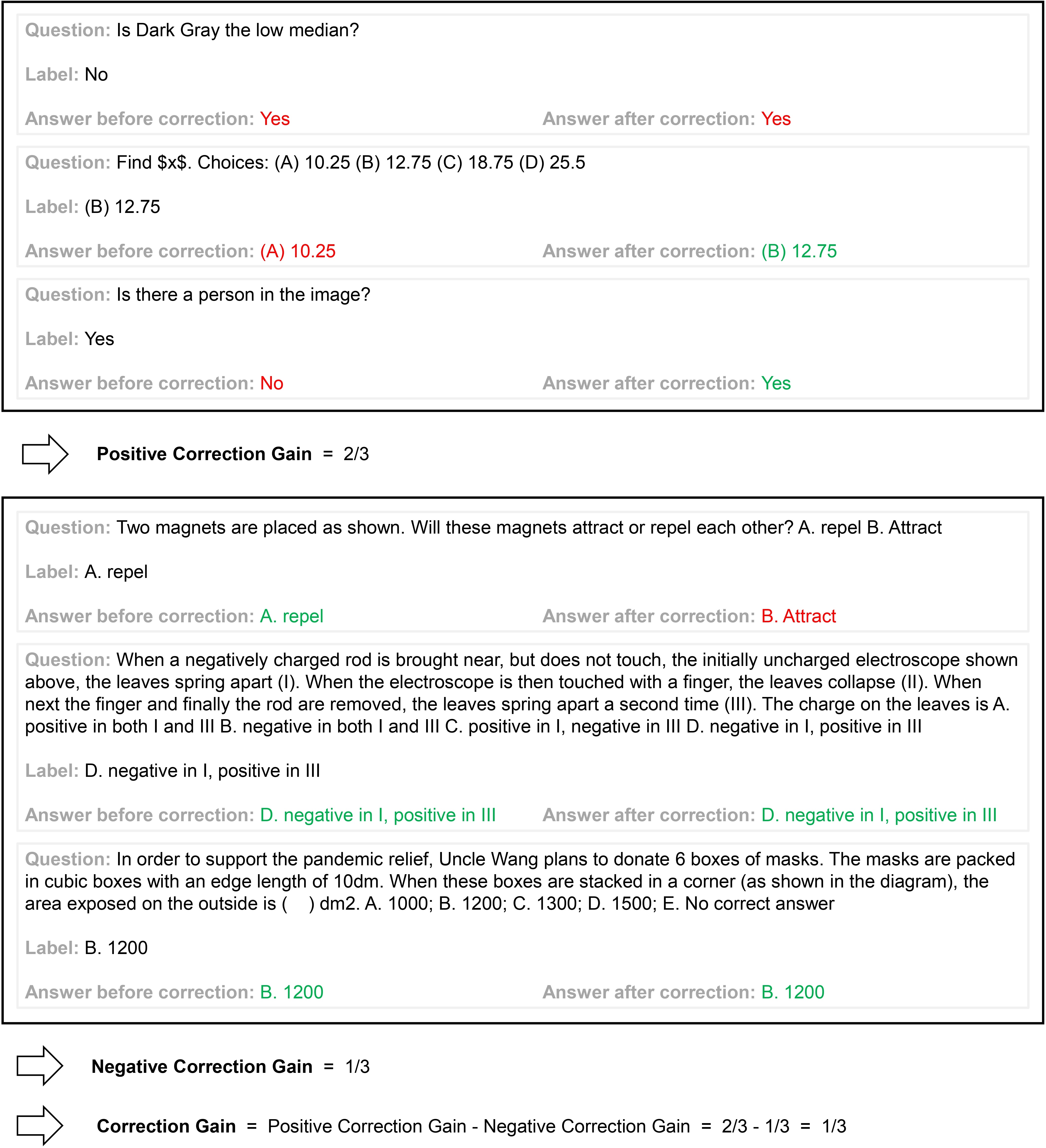}
    \caption{Calculation of correction gain.}
    \label{fig:cg}
    \end{minipage}
    \hfill
    \begin{minipage}{0.48\linewidth} %
        \centering
\includegraphics[width=\linewidth]{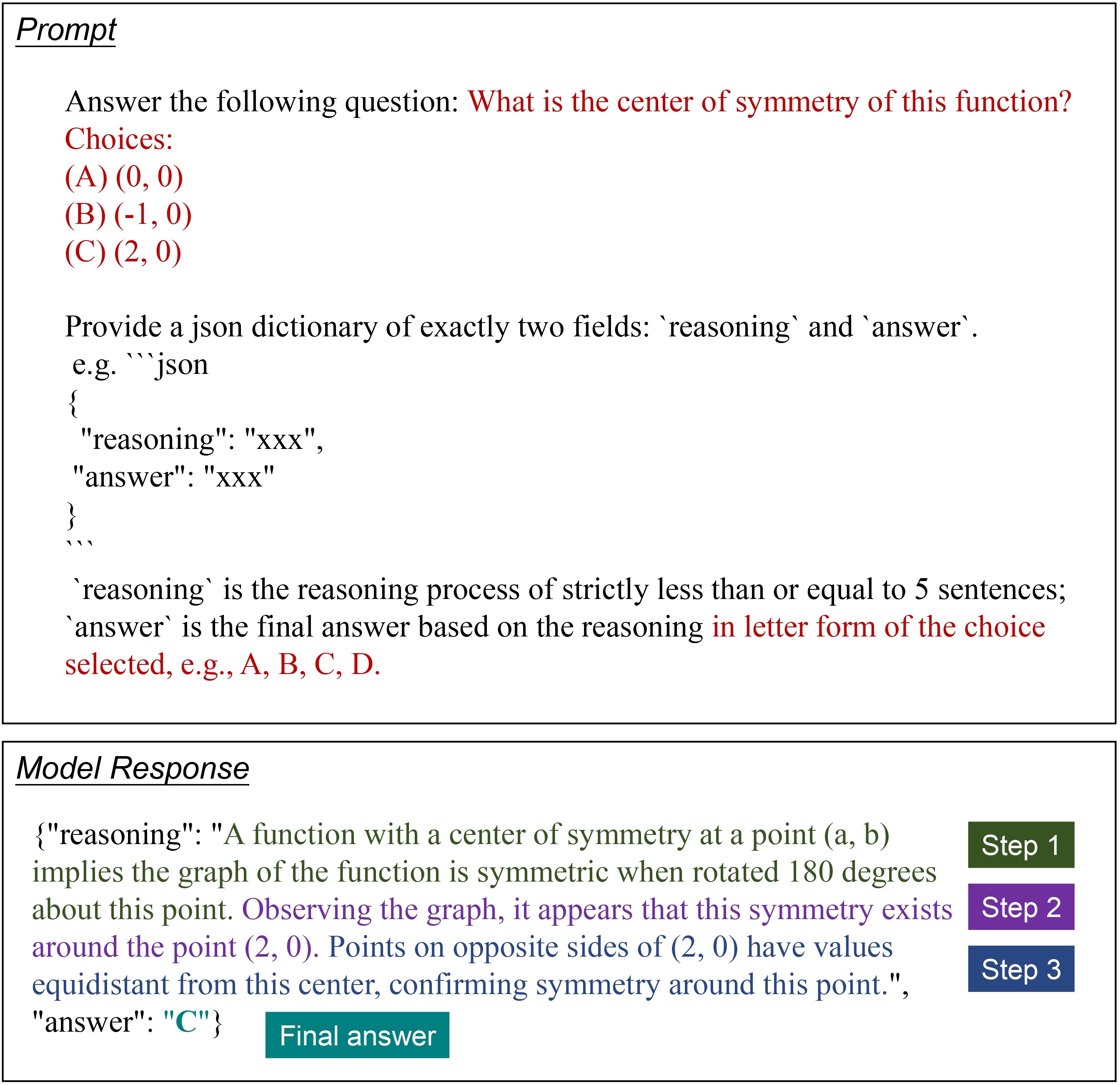}
    \caption{Prompt template for collecting model responses and an example response. The {\color{darkred}highlighted} fields will be filled with the information according to each data point. The ``reasoning'' field in the model response will be split into multiple reasoning steps using \texttt{nltk.tokenize.sent\_tokenize}.}
    \label{fig:response_collection}
    \end{minipage}
\end{figure*}

\begin{figure*}
    \centering
\includegraphics[width=\linewidth]{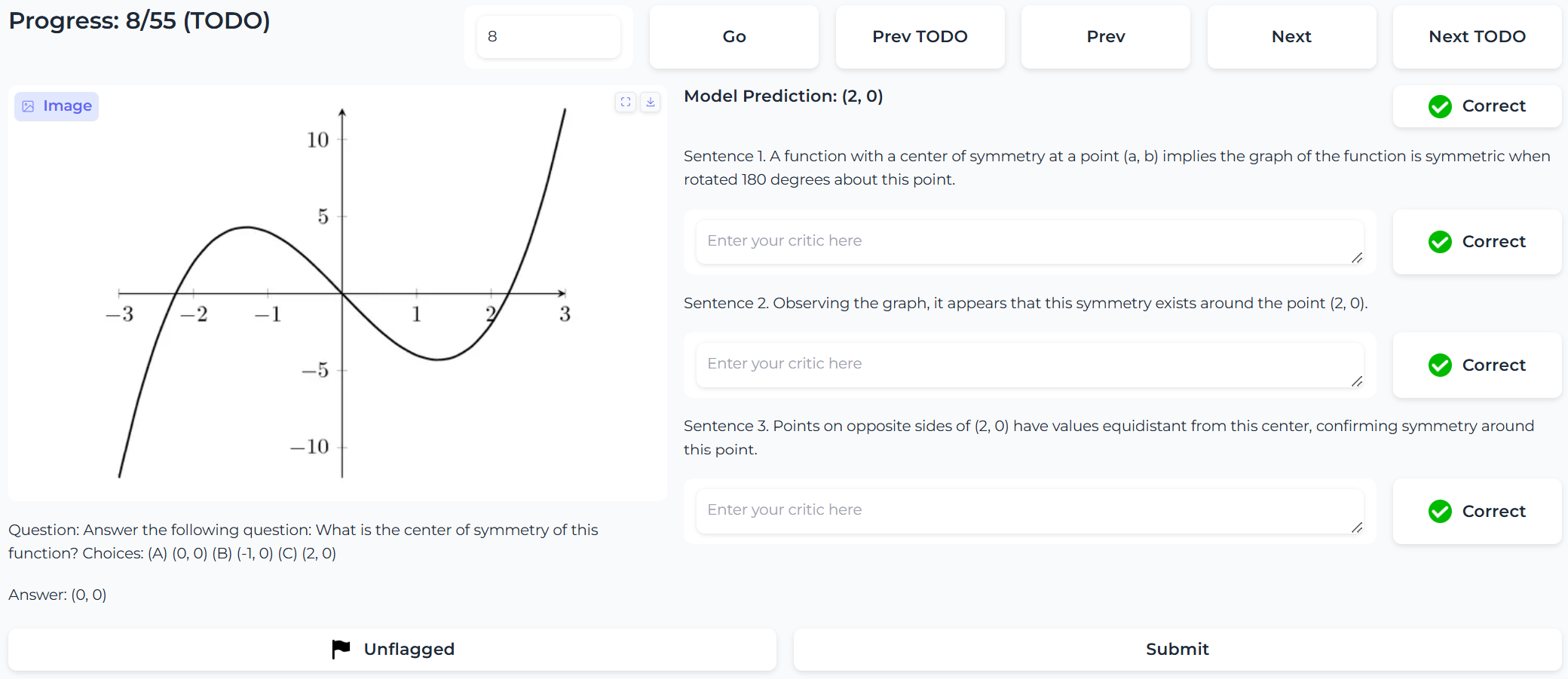}
    \caption{Interface of critique annotation.}
    \label{fig:critique_collection_ui}
\end{figure*}

\begin{figure*}
    \centering
\includegraphics[width=\linewidth]{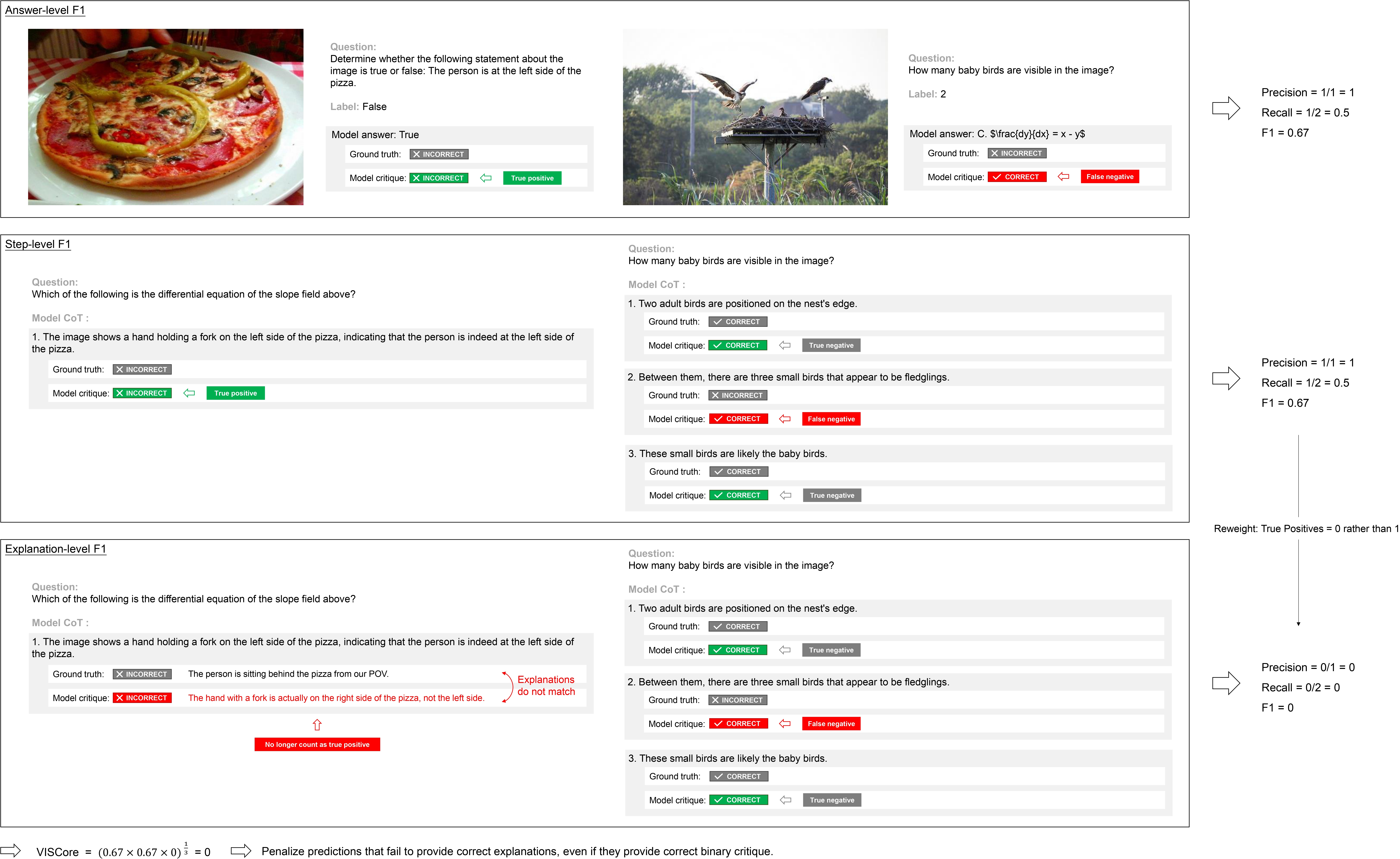}
    \caption{Calculation of \ourscore{}.}
    \label{fig:viscore}
\end{figure*}

In this section, we present additional experimental results. Figure \ref{fig:correction_full} is a more complete version of Figure 9 in the main content, showing the correction performance given model-generated or human-generated critiques with different granularity. We further report the detailed critique performance of each model at different granularity and categories in Table \ref{tab:exp_critique}, and include a few additional models like LLaVA-v1.5.

To measure the importance of CoT to our framework, we perform ablation study by evaluating the critique and correction performance without CoT in the original response. As shown below, LVLMs can effectively perform critique and correction bringing positive correction gains without CoT. However, CoT brings better performance especially for strong models like GPT-4o.
\begin{table}[!h]
    \centering
\setlength{\tabcolsep}{3pt}
\begin{tabular}{l|rr|rr}
\hline
\multirow{2}{*}{Model} & \multicolumn{2}{c|}{Critique} & \multicolumn{2}{c}{Correction} \\
& {\small w. CoT} & {\small w/o CoT} & {\small w. CoT} & {\small w/o CoT} \\
\hline
InternVL2-8B & 37.1 & \textbf{47.3} & \textbf{5.4} & 5.1 \\
Claude-3.5-Sonnet & \textbf{61.8} & 60.3 & \textbf{25.6} & \textbf{25.6} \\
GPT-4o & \textbf{63.0} & 59.0 & \textbf{28.8} & 26.4 \\
\hline
\end{tabular}
    \caption{Additional experiments for ablations of CoT.}
    \label{tab:my_label}
\end{table}

\section{Details on \ourscritic{}}
\label{sec:appendix_ours}

The algorithm for our proposed \ourscritic{} method is in Alg. \ref{alg:ours}. We further show an example of critique generated by \ourscritic{} in Figure \ref{fig:example_ours}.

\begin{algorithm}
\caption{\ourscritic{} algorithm}
\label{alg:ours}
\begin{algorithmic}[1] %
\Require LVLM, image $I$, question $q$, model response to be critiqued $(\mathbf{T}_\mathcal{M}, a_\mathcal{M})$
\State $\mathcal{Q} = \{\}$
\For{$i = 1$ \textbf{to} $N$}
    \State $\mathcal{Q}_i = RaiseQuestion(\text{LVLM}, I, T_i)$ %
    \State $\mathcal{Q} = \mathcal{Q} \cup \mathcal{Q}_i$
\EndFor
\State $\mathcal{Q} = \mathcal{Q} \cup \{q\}$
\State $\mathcal{A} = \{\}$
\For{$q'$ \textbf{in} $\mathcal{Q}$}
    \State $a' = AnswerQuestion(\text{LVLM}, I, q')$
    \State $\mathcal{A} = \mathcal{A} \cup \{a'\}$
\EndFor
\State \Return $Critique\left(\text{LVLM}, I, q, (\mathbf{T}_\mathcal{M}, a_\mathcal{M}), \mathcal{Q}, \mathcal{A}\right)$
\end{algorithmic}
\end{algorithm}

\noindent\textbf{Error cases of \ourscritic{}.} We further evaluate the error cases of \ourscritic{} and identify two typical scenarios: (1) \uline{Errors in reasoning critique}. \ourscritic{} is less effective in critiquing reasoning, as these often go beyond verifying atomic information against the image. This is 
reflected in the smaller performance boosts in reasoning tasks in Table 4.
(2) \uline{Referral expression error in atomic information extraction}. In an example involving four images, the reasoning discusses a blue car \textit{in the first image}, but the extracted question refers to the car as \textit{the blue car} without specifying the first image, resulting in critique errors.

\section{Limitations}

The main limitation of this work is its task setting, which constrains the CoT to at most five reasoning steps, potentially affecting its generalization to complex reasoning or multi-turn interactions.  While this work provides a valuable initial investigation into LVLM critique and correction, this limitation highlights opportunities for future research.

\begin{table*}[!htbp]
    \centering
    \small
\begin{tabular}{l|r|rrr|rrr|rrr}
    \toprule
& \multirow{2}{*}{\ourscore{}} & \multicolumn{3}{c|}{Total} & \multicolumn{3}{c|}{Reasoning} & \multicolumn{3}{c}{Perception} \\
& & \small Answer & \small Thought & \small Expl. & \small Answer & \small Step & \small Expl. & \small Answer & \small Step & \small Expl. \\
\midrule
\rowcolor[RGB]{240,240,240}Random & - & 37.91 & 32.02 & - & 39.44 & 33.10 & - & 36.03 & 30.54 & - \\
\midrule
\multicolumn{11}{c}{\textit{Tiny-size ($<$3B) Open LVLMs}} \\
\midrule
Qwen2-VL-2B & 9.76 & 30.7 & 23.0 & 1.3 & 29.5 & 24.3 & 1.7 & 31.9 & 21.3 & 0.8 \\
InternVL2-2B & \textbf{13.96} & \textbf{36.1} & \textbf{28.2} & \textbf{2.7} & \textbf{37.3} & \textbf{29.3} & \textbf{3.7} & \textbf{34.8} & \textbf{26.8} & \textbf{1.5} \\
\midrule
\multicolumn{11}{c}{\textit{Small-size ($\sim$7B) Open LVLMs}} \\
\midrule
DeepSeek-VL-7B & 7.53 & 21.8 & 15.7 & 1.2 & 18.7 & 14.4 & 1.6 & 24.9 & 17.5 & 0.8 \\
LLaVA-v1.5-7B & 6.51 & 11.9 & 13.4 & 1.7 & 12.7 & 13.5 & 2.0 & 10.9 & 13.3 & 1.4 \\
LLaVA-v1.6-7B & 21.80 & \textbf{44.6} & \textbf{33.6} & 6.9 & \textbf{46.8} & 34.6 & 7.8 & \textbf{41.3} & \textbf{31.9} & 5.6 \\
LLaVA-v1.6-Vicuna-7B & 11.45 & 27.7 & 24.1 & 2.3 & 25.8 & 24.3 & 2.2 & 29.9 & 23.8 & 2.4 \\
LLaVA-OV-7B & 7.53 & 14.5 & 14.9 & 2.0 & 16.8 & 15.7 & 2.2 & 11.7 & 13.8 & 1.6 \\
Qwen-VL-7B & 12.69 & 33.1 & 26.4 & 2.3 & 31.2 & 26.7 & 2.8 & 35.1 & 26.0 & 1.8 \\
Qwen2-VL-7B & 21.71 & 43.0 & 30.6 & 7.8 & 44.5 & 30.4 & 7.3 & 41.1 & 31.0 & \textbf{8.4} \\
Molmo-7B & 13.43 & 35.5 & 22.0 & 3.1 & 35.7 & 23.6 & 4.1 & 35.3 & 19.8 & 1.6 \\
InternVL2-8B & \textbf{23.33} & 37.1 & 31.1 & \textbf{11.0} & 44.2 & \textbf{37.4} & \textbf{14.1} & 26.1 & 19.7 & 5.6 \\
MiniCPM-V2.6 (8B) & 13.07 & 27.9 & 18.2 & 4.4 & 32.0 & 21.7 & 5.5 & 22.4 & 12.4 & 2.7 \\
\midrule
\multicolumn{11}{c}{\textit{Medium-size (10$\sim$70B) Open LVLMs}} \\
\midrule
Llama-3.2-11B & 11.44 & 29.4 & 21.1 & 2.4 & 31.4 & 23.0 & 3.3 & 26.9 & 17.6 & 0.8 \\
LLaVA-v1.6-13B & 21.02 & 40.2 & \textbf{32.8} & 7.1 & 43.6 & 36.5 & 7.6 & \textbf{34.4} & \textbf{25.9} & 6.1 \\
InternVL2-26B & 25.20 & 39.4 & 30.2 & 13.4 & \textbf{48.7} & 36.8 & 16.9 & 24.5 & 18.5 & 7.2 \\
LLaVA-v1.6-34B & 11.05 & 23.6 & 14.3 & 4.0 & 29.4 & 17.4 & 4.2 & 15.3 & 9.5 & 3.6 \\
InternVL2-40B & \textbf{28.48} & \textbf{41.6} & 31.4 & \textbf{17.7} & 47.8 & \textbf{37.4} & \textbf{20.7} & 32.0 & 20.7 & \textbf{12.4} \\
\midrule
\multicolumn{11}{c}{\textit{Large-size ($>$70B) Open LVLMs}} \\
\midrule
LLaVA-OV-72B & 35.27 & 47.1 & 42.0 & 22.2 & 49.3 & 44.7 & 23.2 & \textbf{43.9} & \textbf{37.3} & \textbf{20.5} \\
Qwen2-VL-72B & \textbf{37.44} & 49.2 & 41.9 & \textbf{25.5} & 56.3 & 49.0 & \textbf{30.8} & 38.0 & 28.6 & 15.5 \\
NVLM-72B & 33.07 & 44.0 & 38.6 & 21.3 & 52.6 & 46.1 & 26.1 & 30.6 & 24.7 & 12.3 \\
Molmo-72B & 35.59 & \textbf{49.4} & 39.8 & 22.9 & \textbf{57.5} & 46.8 & 26.8 & 36.5 & 26.4 & 15.5 \\
InternVL2-76B & 26.38 & 37.7 & 28.6 & 17.0 & 47.3 & 35.6 & 21.2 & 22.9 & 16.2 & 9.6 \\
Llama-3.2-90B & 36.40 & 46.8 & \textbf{42.5} & 24.3 & 55.7 & \textbf{49.1} & 28.8 & 32.9 & 30.5 & 16.1 \\
\midrule
\multicolumn{11}{c}{\textit{Critique LVLMs}} \\
\midrule
Prometheus-Vision-7B & 17.67 & 37.6 & 35.8 & 4.1 & 37.2 & 37.6 & 3.1 & 38.0 & 33.4 & 5.4 \\
LLaVA-Critic-7B & 20.02 & 32.0 & 28.7 & 8.8 & 36.3 & 30.3 & 9.5 & 26.3 & 26.3 & 7.7 \\
Prometheus-Vision-13B & 19.32 & 38.0 & 37.8 & 5.0 & 42.0 & 40.1 & 4.6 & 33.1 & 34.5 & 5.6 \\
LLaVA-Critic-72B & \textbf{42.60} & \textbf{53.9} & \textbf{50.9} & \textbf{28.2} & \textbf{56.3} & \textbf{54.8} & \textbf{28.6} & \textbf{50.5} & \textbf{44.8} & \textbf{27.4} \\
\midrule
\multicolumn{11}{c}{\textit{Proprietary LVLMs}} \\
\midrule
Gemini-1.5-Pro & 45.01 & 55.6 & 51.2 & 32.0 & 58.8 & 55.0 & 35.7 & 51.2 & 45.2 & 26.3 \\
Claude-3.5-Sonnet & 51.28 & 61.8 & \textbf{58.1} & 37.6 & \textbf{65.4} & \textbf{61.6} & 43.5 & 57.4 & \textbf{53.2} & 29.5 \\
GPT-4o & \textbf{52.36} & \textbf{63.0} & 57.2 & \textbf{39.8} & 65.2 & 61.4 & \textbf{43.5} & \textbf{60.1} & 50.3 & \textbf{34.0} \\
\midrule
\rowcolor[RGB]{240,240,240}Human$^*$ & 86.47 & 100.0 & 90.6 & 71.4 & 100.0 & 90.6 & 66.8 & 100.0 & 90.6 & 76.4 \\
\bottomrule
    \end{tabular}
    \caption{\textbf{Detailed critique performance \ours{}.} In addition to the overall \ourscore{} score, we also report the critique F1 at three granularity: answer-level F1, step-level F1, and explanation-level F1. We also report the scores for the two categories, reasoning and perception respectively. We also include a few additional models compared to Table 3 in the main content.}
    \label{tab:exp_critique}
\end{table*}

\begin{figure*}
\centering
\fontsize{7}{9}\selectfont
\begin{tabularx}{\textwidth}{X}
\hline
You task is to evaluate a **critique** of reasoning.\\
\\
You are given a question about an image, an incorrect chain-of-thought trying to answer the question, and a **critique** that explains why the \texttt{\color{darkred}STEP\_NUM} step of the chain-of-thought is incorrect. You are required to focus on the \texttt{\color{darkred}STEP\_NUM} step, and analyzes whether the critique correctly identifies the source of error.\\
\\
For reference, you will be provided with the ground truth critique for each individual step. Evaluate the critique by comparing against the ground truth. Focus on whether the critique accurately identifies the **core mistake**. If the critique addresses the core error and does not contain factual or logical error, minor deviations or omissions in reasoning or explanation should be considered as correct.\\
\\
Think step by step, then provide your judgment. Your response should end with either:\\
\\
\# Judgment: the critique is correct.\\
\\
or\\
\\
\# Judgment: the critique is incorrect.\\
\\
\texttt{\color{NavyBlue}... In-context examples ...}\\
\\
\# Question: \texttt{\color{darkred}QUESTION}\\
\\
\# Chain-of-thought: \texttt{\color{darkred}CoT}\\
\hline
\end{tabularx}
    \vspace{-1.2em}
\caption{\textbf{Prompt for LLM-assisted evaluation of critique explanation.} The {\color{darkred}highlighted} fields are to be filled according to each data point. We omit the {\color{NavyBlue}in-context examples} here.}
\label{fig:prompt_gpt_eval}
\end{figure*}

\begin{figure*}
\centering
\fontsize{7}{9}\selectfont
\begin{tabularx}{\textwidth}{X}
\hline
You are given an image, a question about the image, a reasoning process involving multiple steps, and a final answer. Evaluate the accuracy of each reasoning step and the final answer. For each reasoning step, evaluate whether it is correct or incorrect. If it is incorrect, briefly explain why. Then, evaluate the final answer as correct or incorrect.\\
\\
Your response should be a json in the following format:\\
\textasciigrave{}\textasciigrave{}\textasciigrave{}json\\
\{\\
  \texttt{"}step\_1\texttt{"}: \{\texttt{"}correctness\texttt{"}: true or false, \texttt{"}explanation\texttt{"}: \texttt{"}Write your explanation here\texttt{"}\},\\
  ...,\\
  \texttt{"}answer\_correctness\texttt{"}: true or false\\
\}\\
\textasciigrave{}\textasciigrave{}\textasciigrave{}\\
\\
---\\
\\
Instructions for evaluating reasoning steps:\\
* For each reasoning step, evaluate whether it is correct or incorrect based on the accuracy of the factual information and logical calculations it contains.\\
* Evaluate each step in isolation.\\
* You do not need to evaluate the importance of the step in achieving the correct final answer; focus solely on the correctness within that step itself.\\
\\
---\\
\\
\# Question: \texttt{\color{darkred}QUESTION}\\
\\
\# Reasoning: \texttt{\color{darkred}CoT}\\
\\
\# Answer: \texttt{\color{darkred}ANSWER}\\
\hline
\end{tabularx}
    \vspace{-1.2em}
\caption{\textbf{Prompt for critique task.} The {\color{darkred}highlighted} fields are to be filled according to each data point.}
\label{fig:prompt_critique}
\end{figure*}

\begin{figure*}
\centering
\fontsize{7}{9}\selectfont
\begin{tabularx}{\textwidth}{X}
\hline
You are given an image, a question about the image, a multi-step reasoning process leading to an answer, and the critique for each reasoning step. Based on this information, think step by step and provide the correct answer. Your response should end with a json dictionary as follows:\\
\textasciigrave{}\textasciigrave{}\textasciigrave{}json\\
\{\texttt{"}answer\texttt{"}: ANSWER\}\\
\textasciigrave{}\textasciigrave{}\textasciigrave{}\\
ANSWER should be \texttt{\color{darkred}ANSWER\_FORMAT}.\\
\\
\# Question: \texttt{\color{darkred}QUESTION}\\
\\
\# Reasoning:\\
1. \texttt{\color{darkred}REASONING\_STEP\_1}\\
~~- Critique: \texttt{\color{darkred}CRITIQUE\_FOR\_STEP\_1}\\
\texttt{\color{darkred}...}\\
The final answer is: \texttt{\color{darkred}MODEL\_RESPONSE\_ANSWER}\\
~~- Critique: \texttt{\color{darkred}CRITIQUE\_FOR\_ANSWER}\\
\hline
\end{tabularx}
    \vspace{-1.2em}
\caption{\textbf{Prompt for correction task.} The {\color{darkred}highlighted} fields are to be filled according to each data point.}
\label{fig:prompt_correction}
\end{figure*}

\begin{figure*}
    \centering
\includegraphics[width=0.9\linewidth]{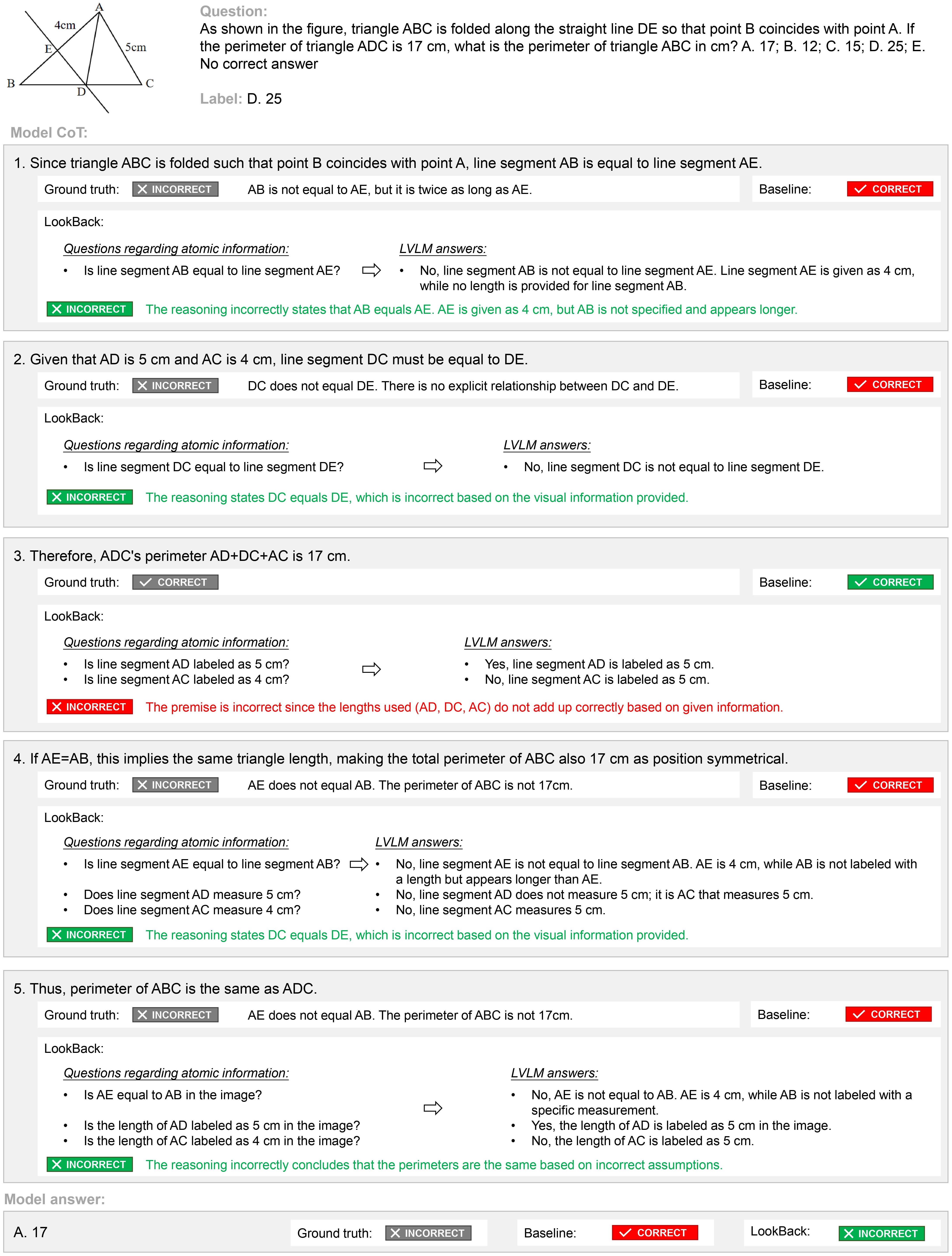}
    \caption{Example of \ourscritic{}.}
    \label{fig:example_ours}
\end{figure*}

\end{document}